\documentclass[letterpaper]{article} 
\usepackage[]{aaai2026}  
\usepackage{times}  
\usepackage{helvet}  
\usepackage{courier}  
\usepackage[hyphens]{url}  
\usepackage{graphicx} 
\usepackage{natbib}  
\usepackage{caption} 
\usepackage{algorithm}
\usepackage{algorithmic}
\usepackage{todonotes}
\usepackage{enumitem}

\usepackage{listings}
\usepackage{xcolor}
\lstdefinelanguage{GraphQL}{
  morekeywords={query,mutation,subscription,type,fragment,on,input,enum,interface,union,scalar},
  sensitive=true,
  morecomment=[l]{\#},
  morestring=[b]",
}

\usepackage{newfloat}
\usepackage{listings}
\usepackage{makecell} 
\DeclareCaptionStyle{ruled}{labelfont=normalfont,labelsep=colon,strut=off} 
\floatstyle{ruled}
\newfloat{listing}{tb}{lst}{}
\floatname{listing}{Listing}
\title{FinOps Agent - A Use-Case for IT Infrastructure and Cost Optimization}

\author{ \\
   Ngoc Phuoc An Vo\textsuperscript{\rm 1}, Manish Kesarwani\textsuperscript{\rm 2}, Ruchi Mahindru\textsuperscript{\rm 1}, Chandrasekhar Narayanaswami\textsuperscript{\rm 1}
}
\affiliations{
    \textsuperscript{\rm 1}IBM Research, Yorktown Heights, US\\
    \textsuperscript{\rm 2}IBM Research, Bengaluru, India\\

    ngoc.phuoc.an.vo@ibm.com, manishkesarwani@in.ibm.com, rmahindr@us.ibm.com, chandras@us.ibm.com
}

\begin{document}

\maketitle

\begin{abstract}
FinOps (Finance + Operations) represents an operational framework and cultural practice which maximizes cloud business value through collaborative financial accountability across engineering, finance, and business teams. FinOps practitioners face a fundamental challenge: billing data arrives in heterogeneous formats, taxonomies, and metrics from multiple cloud providers and internal systems which eventually lead to synthesizing actionable insights, and making time-sensitive decisions. To address this challenge, we propose leveraging autonomous, goal-driven AI agents for FinOps automation. In this paper, we built a FinOps agent for a typical use-case for IT infrastructure and cost optimization. We built a system simulating a realistic end-to-end industry process starting with retrieving data from various sources to consolidating and analyzing the data to generate recommendations for optimization. We defined a set of metrics to evaluate our agent using several open-source and close-source language models and it shows that the agent was able to understand, plan, and execute tasks as well as an actual FinOps practitioner.
\end{abstract}


\section{Introduction}
Cloud infrastructure has become central to the operation of many modern enterprises. The resulting ease of deployment and flexibility of scaling infrastructure as needed, coupled with dynamic pricing models from hyperscalers has created unprecedented complexity in IT cost management.
This has given rise to the discipline of FinOps, which drives strategic business value creation, not just operational cost savings.

FinOps (Finance + Operations) represents an operational framework and cultural practice that maximizes cloud business value through collaborative financial accountability across engineering, finance, and business teams. The framework operates through three iterative phases: Inform (Visibility \& Allocation), Optimize (Rates \& Usage), and Operate (Continuous Improvement \& Usage).
Importantly, success in FinOps hinges on making continuous adaptive changes using real-time insights with data gathered from Application Performance Management (APM), Application Resource Management (ARM), and Finance (Cloud Cost Management)~\cite{cloud_finops_2nd}.

The FinOps Foundation's 2025 State of FinOps report \cite{finops2025} curated data from 861 respondents totaling \$69 billion in public cloud spend, with 31\% currently spending more than \$50 million annually in public cloud, 20\% spending more than \$100 million, and over 20 organizations spending \$1 billion annually. Not surprisingly, AI spending management has emerged as a critical concern, with 63\% of respondents now managing AI costs — up from 31\% the previous year — signaling a fundamental shift in FinOps priorities. Similar to the observations made in the 2024 State of FinOps Report, workload optimization and waste reduction remained key priorities across the board. 
Management of cloud discount programs (such as Savings Plans and Reserved Instances) and accurate forecasting of spend continued to be top concerns.


FinOps practitioners face a fundamental challenge: relevant data arrives in heterogeneous formats, taxonomies, and metrics from multiple cloud providers, other IT vendors, and internal systems. Synthesizing actionable insights from these disparate sources requires understanding complex relationships between IT bills, resource utilization patterns, observability data, IT resource types,  cost and performance KPIs, hyperscaler pricing constructs, organizational hierarchies, etc. Time-sensitive decision-making compounds this complexity, as delayed insights can result in significant financial impact. We note that the disciplines of FinOps and SRE are intertwined due to the interplay between cost and performance. 

In order to assist practitioners, the FinOps Foundation \cite{finopsbenchkpis} provides benchmarks that compare cloud financial performance across organizations and departments, focusing on KPIs such as resource utilization efficiency, contract coverage, and cost apportionment. These benchmarks help assess cloud efficiency by evaluating internal and external metrics, fostering structured, collaborative approaches to cloud optimization.

Several tools, platforms, and standards have also emerged in this space. Two standards have emerged - FOCUS (FinOps Open Cost and Usage Specification) for providers to offer billing data in a consistent format~\cite{focus}, and TBM (Technology Business Management) \cite{tbmspec} - ``To promote alignment between IT, Finance, and Business Unit leaders, TBM provides a standard taxonomy to describe cost sources, technologies, IT resources (IT towers), applications, and services. Just as businesses rely on generally accepted accounting principles (or GAAP) to drive standard practices for financial reporting — and thus comparability between financial statements — the TBM taxonomy provides a generally accepted way of categorizing and reporting IT costs and other metrics." 

The hyperscalers now provide tools for monitoring costs such as AWS Cost Explorer, GCP Cost Explorer, Microsoft Cost Management, and Oracle's FinOps Hub. Independent software vendors such as  CloudZero, Apptio, ServiceNow, ProsperOps, Nicus, Magic Orange, etc., also provide tools to monitor and manage IT spending. These tools typically ingest data from multiple sources and provide dashboards, provide forecasts, identify and flag trends, categorize expenses, compute chargebacks, plan capacity, etc.

\paragraph{\textit{Introduction for Agentic AI }}
Recent advances in Agentic AI offer a transformative approach to these challenges. Agentic AI introduces autonomous agents capable of goal-directed behavior, leveraging strategic planning, and adaptive learning \cite{yao2023react, wang2024survey}, rather than just reacting literally to direct user prompts. These agents leverage the natural language understanding of large language models while adding critical capabilities: they can formulate multi-step plans,  include reflection loops and learn from feedback, execute actions through tool integration,  retrieve data from various systems, have memory to keep state over time, and adapt strategies based on currently observed conditions. 

The above capabilities can enhance FinOps by allowing proactive optimization and assisting reactive analysis by providing detailed explanations, both of which are essential to practitioners. In the realm of proactive optimization, an agent could continuously monitor cloud spending, autonomously investigate irregularities, correlate costs with business events, and execute approved optimizations.  For example, when detecting an unusual spike in compute costs, an agent can autonomously trace it to specific services, analyze recent deployments, identify inefficient resource configurations, evaluate multiple optimization strategies considering both cost and performance impacts, and either present detailed recommendations with predicted savings or implement approved corrective actions, all while maintaining the audit trail required for financial governance. The methodology used by Agentic AI allows it to naturally show explorations and decisions step by step and provide explainability in a graphical manner. 

While existing agent frameworks like ReAct \cite{react} have shown promise in general task automation, applying agentic AI to FinOps requires addressing domain-specific challenges, including understanding complex pricing models, maintaining financial compliance, and balancing multiple stakeholder objectives.

\paragraph{\textbf{\textit{AI Agents for FinOps}}} To further address the needs of FinOps practitioners, we have embarked on the path of creating an AI-based FinOps Agent that can help in various day to day tasks they encounter.  Our agent interprets natural language user queries to perform complex analytical tasks on IT data obtained from multiple sources,  and when necessary interacts with other agents or hands off certain tasks to other agents, such as an SRE agent. 

Major cloud providers are also integrating such agents into their platforms to enhance FinOps practices and, in turn, enable users to manage cloud spend across their portfolios \cite{finopsx}. For example, AWS offers Cost Explorer \cite{awsCost} and Cost Anomaly Detection \cite{awsAD}, which FinOps practitioners can use to recommend ideal compute resources to balance performance and cost as well as identify unexpected spending hikes. We expect this area to have explosive growth in the coming years. 

\paragraph{\textit{Our Contributions}}
In this paper, we introduce a FinOps agent for a specific yet well representative use-case for IT infrastructure and cost optimization. Our contribution consists of the following: 1) a unified schema from various data sources using GraphQL, 2) an NL2GraphQL (Natural Language to GraphQL) layer assisting access to the unified data schema, 3) a multi-agent FinOps agent with tool calling capability for data retrieval as well as consolidating and recommending optimization, and 4) a set of metrics to evaluate the FinOps agent with several open-source and close-source LLMs.

\section{Background and Related Works}



\paragraph{\textit{GraphQL and its Usages}} GraphQL \cite{graphqlspecs} is a query language for APIs and a runtime for executing those queries, initially developed by Facebook in 2012. It was created to overcome the limitations of traditional RESTful APIs, particularly the problems of over-fetching or under-fetching data that clients faced. Unlike REST, which often requires multiple endpoints to gather related data, GraphQL enables clients to specify exactly what data they need in a single request, optimizing network usage and improving app performance. Facebook open-sourced GraphQL in 2015, 
catalyzing its wide adoption beyond the social media giant.

Comparative studies between REST and GraphQL APIs have identified several advantages of GraphQL. The study \cite{brito2019migrating} demonstrated that GraphQL effectively reduces client–server interactions and decreases JSON payload sizes, while \cite{brito2020rest} reported that GraphQL queries are generally easier to implement. Additional investigations \cite{seabra2019rest, mikula2020comparison} further substantiate these findings by examining various performance and usability benefits associated with GraphQL.

Beyond its comparative advantages over REST, recent research has expanded toward testing methodologies for GraphQL queries \cite{belhadi2024random} and systematic mapping analyses \cite{mera2023graphql}. GraphQL has also emerged as a critical component in both academic and industrial contexts and businesses \cite{theirstack}, demonstrating strong potential for facilitating data access and integration across heterogeneous information sources \cite{li2024ontology}. From an industry perspective, its adoption trajectory is expected to continue rising; according to a recent Gartner report, more than 60\% of enterprises are projected to employ GraphQL in production environments by 2027, compared to fewer than 30\% in 2024 \cite{gartner2024}.


\paragraph{\textit{FinOps Agents}} 
Recent research on FinOps agents has progressed from basic automation to sophisticated autonomous systems, though significant gaps remain for production deployment. A significant body of work highlights the development of intelligent, self-improving agents that reduce manual effort in cloud cost management by automating tasks such as real-time resource tracking, cost allocation, and anomaly detection. 

Early FinOps agents focused on automating routine tasks and detecting anomalies. \cite{burke2024improving} demonstrated ML-based agents that reduced false positives in cost anomaly detection by 73\% through continuous learning. \cite{solanke2021cloud} developed real-time resource tracking agents that decreased manual effort by 60\%. The ABACUS framework \cite{deochake2024abacus} advanced this with automated budget analysis and surveillance across distributed cloud accounts. However, these systems operated on predefined rules, requiring human intervention for remediation and lacking adaptability to dynamic pricing models.

Second-generation agents incorporated predictive analytics for proactive management. \cite{nawrocki2024finops} proposed self-improving agents using reinforcement learning that achieved 31\% cost reduction through anticipatory resource provisioning. Multi-time series forecasting techniques improved budget accuracy by 42\%, with agents predicting usage patterns across multiple resource types simultaneously \cite{chen2024multivariate}. Despite these advances, predictive agents failed catastrophically during unexpected events, with errors exceeding 200\% during traffic spikes, limiting their reliability in production environments \cite{anderson2024limitations}.

Large language models enabled natural language interfaces for FinOps. \cite{wang2024llmfinops} fine-tuned LLMs on billing data, achieving 78\% accuracy in cost queries but exhibiting 19\% hallucination rates in optimization recommendations. Conversational agents allowed policy specification in natural language with 89\% interpretation accuracy \cite{kim2024conversational}. While improving accessibility, LLM-based agents struggled with numerical reasoning and showed inconsistent behavior when scenarios were rephrased, highlighting the gap between language understanding and financial optimization.

Real-world deployments revealed critical gaps. Studies of enterprise deployments found only 20\% achieved full automation goals, with common failures including integration complexity (average 47 APIs required), governance violations (31\% of optimizations reversed for compliance), and edge case handling (23\% failure rate on unseen scenarios) \cite{johnson2024production}. Successful deployments required gradual automation, extensive testing, and maintained human oversight with rollback capabilities \cite{davis2024lessons}.


Another important research thrust involves the integration of FinOps agents within larger AI and operational ecosystems. Studies report how FinOps intelligent agents contribute to accelerating financial workflows and reducing errors by up to 94\%, thereby transforming cloud spending governance for AI-native enterprises \cite{cprime2025}. Research also covers multi-time series forecasting for cloud resource optimization, showing that machine learning techniques can predict long-term usage trends to improve budgeting and resource allocation. 

\section{GraphQL for FinOps Agent}

FinOps data is inherently fragmented across multiple cloud providers and vendor platforms, each exposing distinct metrics, taxonomies and access patterns.  In our use case, critical signals for cost-optimization, for example resource utilization from Turbonomic \cite{IBMTurbonomic} and spending anomalies from Apptio \cite{IBMApptio}, reside in separate systems. GraphQL allows the FinOps agent to federate such sources through a single schema-driven endpoint, enabling it to retrieve only the needed fields and eliminate the overhead of multiple REST calls. This reduces over-fetching, lowers latency, and supports time-sensitive FinOps decisions in dynamic cloud environments. Large financial institutions such as National Australia Bank and Goldman Sachs have also adopted GraphQL as a unified query interface to overcome data silos and inconsistent APIs \cite{NordicAPIs2024}. So, the ideas from this paper should apply equally well to platforms for Banking, Trading and other Financial Services.

GraphQL’s introspection capabilities expose a strongly typed schema to the client, which the agent leverages as structured domain knowledge. By grounding its reasoning on this schema, the agent systematically composes queries for tasks such as detecting anomalous spending events or recommending rightsizing actions, rather than relying on heuristic patterns or fixed templates. This schema-based reasoning aligns with recent advances in NL2GraphQL research~\cite{kesarwani2024graphql, gupta2025schema, sonthalia2025robust}, which demonstrate that structured schema representations improve query accuracy through LLM-driven generation with dynamic in-context learning. Collectively, these capabilities unify access to multi-vendor data, enable precise and efficient data extraction, and establish the architectural foundation for the agent’s natural-language-to-query translation process.

Figure~\ref{fig:nl2graphql} (Appendix 2) illustrates the overall architecture linking the heterogeneous data sources, the unified GraphQL schema, and the NL2GraphQL query generation modules with the FinOps agent’s reasoning loop. The subsequent subsections describe each component in detail.

\subsection{Building a Unified GraphQL Schema}

The unified GraphQL schema serves as the FinOps agent’s access layer, abstracting data from multiple vendor systems, such as Turbonomic and Apptio, through a federated query interface. Although these platforms are independent and maintain distinct data models, the integration is handled within GraphQL resolver functions. These resolvers manage entity reconciliation and field normalization, ensuring that semantically equivalent concepts (e.g., “application” in Turbonomic and “service” in Apptio) are mapped correctly. The user interacts only with the unified schema; the cross-vendor resolution occurs transparently at runtime.

\paragraph{\textit{Schema Definition}}  
The deployed schema exposes six core query endpoints corresponding to the agent’s data retrieval tools. The complete GraphQL schema is shown below. \\

\begin{lstlisting}[language=GraphQL, caption={Unified GraphQL Schema for FinOps Agent}]
# ---------------------------
# Turbonomic Endpoints
# ---------------------------
type Entity {
  id: Int!
  name: String!
  description: String
  cost: Float
  user_id: String!
}

type Action {
  id: ID!
  name: String
  actionType: String
  category: String
  severity: String
  risk: String
  target: String
  costImpact: Float
  businessCriticality: String
}

type Query {
  # Application Discovery
  get_applications_names: [String]
  get_entities(application_name: String!): [Entity]
  # Optimization Recommendations
  get_actions(entity_name: String): [Action]
}

# ---------------------------
# Apptio Endpoints
# ---------------------------
type SpendingAnomaly {
  id: ID!
  application: String
  anomalyType: String
  anomalyValue: Float
  timestamp: String
}

type CommitmentRecommendation {
  id: ID!
  service: String
  currentCoverage: Float
  recommendedCommitment: Float
  potentialSavings: Float
}

type RightsizingRecommendation {
  id: ID!
  resource: String
  currentUtilization: Float
  recommendedSize: String
  estimatedSavings: Float
}

extend type Query {
  # Financial Analysis
  get_spending_anomaly_events(app_name: String): [SpendingAnomaly]
  get_commitment_recommendations: [CommitmentRecommendation]

  # Optimization Recommendations
  get_rightsizing_recommendations(app_name: String): [RightsizingRecommendation]
}
\end{lstlisting}

\paragraph{\textit{Resolver Abstraction}}
Each GraphQL query endpoint corresponds to one or more vendor-specific REST APIs or database queries. The resolver layer performs:  
\begin{enumerate}
    \item Field mapping across heterogeneous schemas (e.g., aligning “entityId” from Turbonomic with “resourceId” from Apptio).  
    \item Name normalization for shared dimensions like application or business unit.  
    \item Response merging and deduplication before returning a unified result to the FinOps agent.  
\end{enumerate}

\paragraph{\textit{Agent Integration}}  
At runtime, the agent invokes these GraphQL tools as part of its reasoning chain. For example, when a user asks, “List cost anomalies for applications with rightsizing actions pending,” the agent issues a federated query that internally joins the Turbonomic and Apptio schemas via their resolvers, without requiring explicit orchestration logic. This seamless abstraction enables efficient, multi-source data access critical for real-time FinOps analysis.

\subsection{NL2GraphQL Query Operations}
On top of the unified schema, we implement an NL2GraphQL layer that enables the FinOps agent to convert natural language query from user into executable GraphQL queries. Rather than relying on fixed templates or static query logic, the agent leverages its LLM-based reasoning capabilities to dynamically compose queries against the schema. The schema structure and tool metadata, such as function names, input arguments, and descriptions, are supplied as part of the agent’s context. For example, the planning agent is aware that the function \texttt{get\_spending\_anomaly\_events} retrieves cost anomalies from Apptio, and \texttt{get\_actions} retrieves optimization recommendations from Turbonomic.

We adopt a ReAct-style agent framework~\cite{yao2022react}, where the agent iteratively reasons over the user’s request, decides an action (e.g., invoking a retrieval tool), and observes the response before continuing its reasoning. When a plan requires data access, the agent produces a structured GraphQL query as the action output. 




\paragraph{\textit{Dynamic Query Composition}}
To generate accurate and executable GraphQL queries, we employ few-shot prompting and in-context learning strategies. The LLM is provided with exemplars of high-level FinOps questions and their corresponding query representations, allowing it to learn the mapping between natural language expressions and schema-constrained operations. We further incorporate dynamic ICL selection, where relevant exemplars are selected based on lexical and structural similarity between the current user input and prior examples, following approaches proposed in \cite{kesarwani2024graphql, gupta2025schema}.

\paragraph{\textit{Validation and Error Correction}}
The self-descriptive nature of GraphQL schemas enables on-the-fly validation. Before execution, the agent checks whether the generated fields and endpoints exist within the current schema context. Invalid or incomplete queries trigger self-correction within the ReAct loop: the agent observes the error, updates its reasoning state, and regenerates the query with corrected structure. This iterative validation reduces runtime errors and ensures schema conformity.

In summary, the NL2GraphQL layer acts as a semantic interface between natural language reasoning and formal data access. By combining LLM-driven reasoning, dynamic in-context adaptation, and schema-grounded validation, it allows the FinOps agent to issue complex, multi-source queries with precision.


\section{FinOps Agent for IT Infrastructure \\and Cost Optimization}


\subsection{The Use-case}
FinOps practitioners require agents with three distinct capability levels. \textit{Reactive agents} enable interactive data exploration through natural language queries and iterative refinements, surpassing the limitations of static dashboards. \textit{Proactive agents} autonomously detect anomalies and usage trends, perform root cause analysis, and recommend remediation actions. \textit{Long-range planning agents} integrate both reactive and proactive capabilities to address strategic optimization challenges. These agents inherently require multi-agent architectures due to the complexity of coordinating analysis across time horizons and data sources.

We focused on long-range planning agents as they encompass the full spectrum of FinOps capabilities. To validate our approach, we selected a representative practitioner challenge: \textit{Review pending resource and cost optimization recommendations to accommodate a new product launch within existing budget constraints.}

Figure \ref{fig:finops_what-why} (Appendix 3) illustrates the complex decision tree this query generates. The agent must navigate multiple investigation paths, with path selection contingent on data from disparate systems. The primary optimization strategies include:

\begin{enumerate}
    \item \textbf{Resource Efficiency} Identify underutilized assets such as idle resources, over-provisioned instances, or workloads running on suboptimal hardware or in costly geographic regions.
    \item \textbf{Spend Analysis} Detect anomalous cost increases and determine their business justification. The agent must correlate spending changes with business KPIs to distinguish legitimate growth from operational inefficiencies requiring SRE intervention.
    \item \textbf{Contract Optimization} Analyze financial commitments, including reserved instances and savings plans. Determine whether existing contracts are fully utilized and identify opportunities for renegotiation or new procurement.
\end{enumerate}

Each path branches into numerous sub-investigations. For instance, resource efficiency analysis might reveal over-provisioned databases, leading to right-sizing recommendations. This, in turn, requires workload pattern analysis and performance impact assessment. The rich interconnection of these decision paths makes this problem ideally suited for agentic AI. Traditional rule-based systems cannot handle the combinatorial complexity, while human analysis cannot scale to the data volumes involved.
This complexity motivated our multi-agent architecture. Each agent specializes in specific aspects of the problem space while collaborating to deliver comprehensive optimization strategies.

\subsection{System Description}
\begin{figure*}[h] 
    \centering 
    \includegraphics[width=0.6\linewidth]{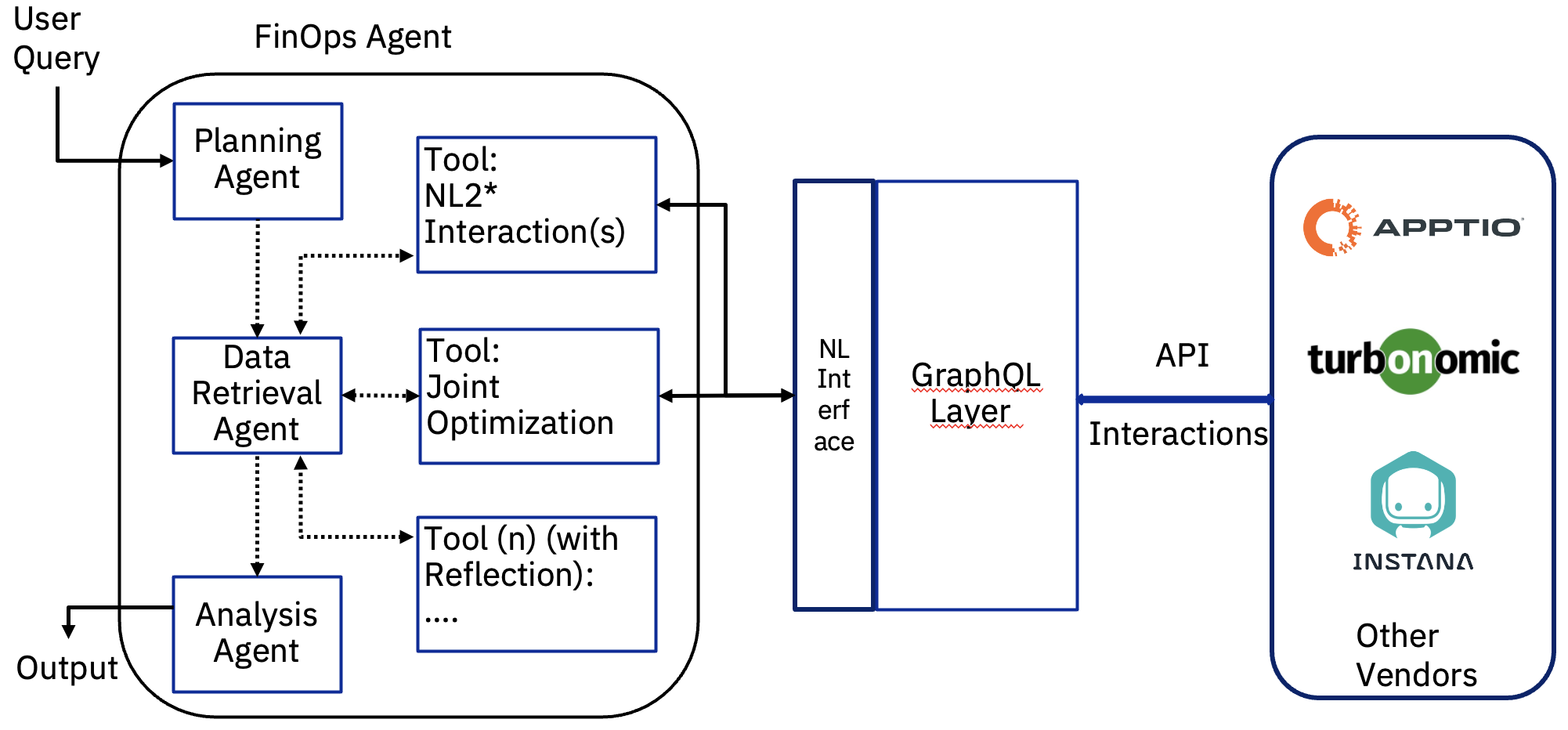}
    \caption{FinOps Agent Architecture.}
    \label{fig:finops_design} 
\end{figure*}

Figure \ref{fig:finops_design} shows the design architecture of our FinOps agent.

\paragraph{\textit{Data Retrieval Tools}} 

The FinOps agent employs six specialized data retrieval tools organized into three functional categories to gather necessary information before performing analysis and generating optimization recommendations.

\begin{enumerate}
    \item Application Discovery
    \begin{itemize}
        \item \texttt{get\_applications\_names()}: Retrieves the catalog of business applications monitored within the infrastructure.
        \item \texttt{get\_entities()}: Fetches infrastructure entities (VMs, containers, storage) associated with specific applications.
    \end{itemize}
    \item Financial Analysis
    \begin{itemize}
        \item \texttt{get\_spending\_anomaly\_events()}: 
        Identifies unusual spending patterns and cost anomalies for specified entities or applications.
        \item \texttt{get\_commitment\_recommendations()}: Analyzes existing reserved instances/savings plan coverage and suggest new commitment purchases.
    \end{itemize}
        \item Optimization Recommendations
    \begin{itemize}
        \item \texttt{get\_actions()}: Obtains optimization recommendations including right-sizing, placement, and scaling actions for selected entities.
        \item \texttt{get\_rightsizing\_recommendations()}: Retrieves right sizing recommendations based on utilization patterns.
    \end{itemize}
\end{enumerate}

\paragraph{\textit{A Multi-agent System}} 
FinOps tasks exhibit inherent complexity. They require coordination across multiple data sources, sophisticated analysis, and domain-specific reasoning. To address this, we implemented a multi-agent architecture using CrewAI\footnote{https://www.crewai.com/}. Our system employs three specialized agents operating within the ReAct framework \cite{yao2022react}. Each agent handles distinct aspects of the optimization workflow.

\begin{itemize}[itemsep=0.1cm]
    \item \textbf{Planning Agent} This agent interprets the user's natural language query. It considers the FinOps knowledge base and available tools to forumate an execution plan. The agent decomposes complex requests into actionable steps. It determines the optimal sequence of operations needed to fulfill the query. Figure \ref{fig:gpt-4o_plan} (Appendix 1) illustrates a sample execution plan generated by \texttt{gpt-4o}. Beyond planning, this agent orchestrates the entire workflow. It monitors progress and adapts the plan based on intermediate results. 
    \item \textbf{Data Retrieval Agent} This agent executes data collection tasks identified in the plan. It invokes appropriate retrieval tools based on data requirements. The agent manages federated queries by translating abstract needs into specific GraphQL queries. It interfaces with multiple vendor APIs (for example, Turbonomic and Apptio).
    \item \textbf{Analysis Agent} This agent synthesizes data from disparate sources into cohesive insights. After data collection completes, it performs cross-system correlation. The agent identifies optimization opportunities based on the consolidated data. It generates actionable recommendations tailored to user constraints and objectives. 
\end{itemize}

This separation of concerns provides key architectural benefits. Each agent can specialize in its domain while maintaining system flexibility. New data sources can be added without modifying planning logic. Analysis strategies can evolve independently of data retrieval mechanisms.


\section{Experiments and Evaluations}

\begin{table*}[htb] 
    \centering
    \begin{tabular}{l|c|c|c|c|c|c|c|c}
        \hline
        \textbf{\thead{Model}} & \textbf{\thead{Execution \\ Time \\ (seconds)}} & \textbf{\thead{Computational \\ Efficiency \\ (Iterations)}} & \textbf{\thead{Planning \\ Accuracy}} & \textbf{\thead{Plan \\ Execution \\ Accuracy}} & \textbf{\thead{Task \\ Completion \\ Rate}} & \textbf{\thead{Tool \\ Recognition \\ Latency}} & \textbf{\thead{Data \\ Consolidation \\ Accuracy}} &  \textbf{\thead{ Recommendation \\ Accuracy}} \\
        \hline
        gpt-4o & 93 & 6 & \textbf{100\%} & 76\% & \textbf{90\%} & \textbf{1} & \textbf{100\%} & \textbf{100\%} \\
        gpt-4o-mini & 93 & 7 & \textbf{100\%} & \textbf{78\%} & 59\% & \textbf{1} & \textbf{100\%} & \textbf{100\%} \\
        Granite-8b & 93 & 2 & \textbf{100\%} & 37\% & 28\% & \textbf{1} & 80\% & 60\% \\
        mistral-large & 302 & 18 & 60\% & 51\% & 55\% & 9 & 80\% & 80\% \\
        llama-405b & 312 & 9 & 35\% & 22\% & 16\% & 5 & 60\% & 60\% \\
        \hline
    \end{tabular}
    \caption{FinOps Agent Performance across LLMs.}
    \label{table:eval}
\end{table*}

\subsection{Experiments}
 We evaluated our FinOps agent using five state-of-the-art language models. \textit{Proprietary Models:} \texttt{gpt-4o}, \texttt{gpt-4o-mini}, \texttt{mistral-large}
\textit{Open-Source Models:} \texttt{Granite-3.1-8b}, \texttt{llama-405b}.

Each model was tested with identical prompts and input queries across 10 independent runs to ensure statistical validity. The temperature was set to 0 for all models to balance consistency with creative problem-solving.
\paragraph{\textit{Ground Truth Creation}} We established comprehensive ground truth data through collaboration with FinOps subject matter experts (SMEs). Our ground truth comprises two components:

\begin{itemize}
    \item \textit{Execution Plan:} We consulted with two FinOps SMEs who helped define the expected execution plan for our use case. This plan specifies the optimal sequence of actions for data retrieval, analysis, and optimization recommendation generation. The experts validated that this sequence represents best practices for addressing budget-constrained infrastructure scaling scenarios.
    
    \item \textit{Execution Data:} We constructed mock datasets simulating real-world data from Turbonomic (infrastructure performance metrics) and Apptio (financial analytics). These datasets include realistic cost anomalies, resource utilization patterns, and optimization opportunities. 
\end{itemize}

\paragraph{\textit{Data Retrieval via NL2GraphQL}} We built a GraphQL schema to extract and join the data derived from two sources, then we built an NL2GraphQL layer on top of it.
We implemented a GraphQL federation layer to unify data access across vendor systems. This architecture includes:
\begin{itemize}
    \item A unified schema abstracting Turbonomic and Apptio data models.
    \item Natural language to GraphQL (NL2GraphQL) translation capabilities.
    \item Query optimization for efficient cross-system joins.
\end{itemize}


\subsection{Evaluation Metrics}
We designed a comprehensive evaluation framework measuring multiple dimensions of agent performance. To evaluate our FinOps agent, we not only rely on the final optimization recommendation accuracy but also defined a set of evaluation metrics to measure different dimensions of the agent (out of 10 runs) as follows:

\begin{itemize}
    \item \textbf{\textit{Performance Metrics}}
    \begin{itemize}
        \item \textit{Execution Time: } Average duration across 10 runs in seconds.
        \item \textit{Computational Efficiency: } Average number of ReAct iterations required to reach completion.
    \end{itemize}
    \item \textit{\textbf{Accuracy Metrics}}
    \begin{itemize}
        \item \textit{Planning Accuracy}: Percentage of runs generating plans that matched the ground truth sequence.
        \item \textit{Plan Execution Accuracy}: Percentage of runs successfully that executed all plan steps without errors
        For example, out of 10 runs, how many runs the plan was executed correctly.
        \item \textit{Data Consolidation Accuracy}: 
        Success rate in correctly merging multi-source data.
        \item \textbf{Recommendation Accuracy: } Percentage of runs that produced valid ServiceNow optimization records.
    \end{itemize}
    \item \textbf{\textit{Tool Interaction Metrics}}
    \begin{itemize}
        \item \textit{Tool Recognition Latency: } Average ReAct iteration before recognizing all available tools. 
        For example, after 1 iteration, \texttt{gpt-4o} can recognize all tools whereas \texttt{mistral-large} can recognize at 9th iteration.
        \item \textit{Task Completion Rate: } Proportion of data retrieval tasks (out of 6) that executed successfully.
    \end{itemize}
\end{itemize}

These metrics enable holistic assessment beyond simple accuracy. They capture the agent's reasoning efficiency, tool utilization capabilities, and practical deployability. By measuring intermediate steps, we can identify failure points and optimization opportunities in the agent pipeline.

\subsection{Results and Analysis}
Table \ref{table:eval} presents comprehensive evaluation results across all models and metrics. \texttt{gpt-4o} and \texttt{gpt-4o-mini} demonstrate superior performance across all metrics. Both achieve perfect planning accuracy (100\%) and data consolidation (100\%). \texttt{gpt-4o} slightly outperforms its mini variant in execution accuracy (76\% vs 78\%) and data retrieval tasks (90\% vs 59\%). Notably, both models recognize tools immediately (1 iteration) and complete tasks efficiently (6-7 ReAct iterations). The consistent 93-second execution time suggests rate limiting rather than computational constraints.
Despite achieving perfect planning accuracy (100\%), \texttt{Granite-8b} struggles with execution (37\% accuracy) and data retrieval (28\% task completion). Its efficiency is noteworthy as it requires only 2 ReAct iterations, but this brevity comes at the cost of incomplete task execution. The model shows degraded performance in complex synthesis tasks (80\% consolidation, 60\% recommendation accuracy). 
These models exhibit fundamental limitations for FinOps tasks. \texttt{mistral-large} requires 9 iterations just to recognize tools and 18 total iterations, resulting in 3x longer execution times (302 seconds). \texttt{llama-405b} performs worst overall with only 35\% planning accuracy and 22\% execution success. Both models fail to reliably complete even basic data retrieval tasks (55\% and 16\% respectively).

Our analysis reveals several critical insights about model performance in FinOps tasks. \textit{Tool recognition emerges as a crucial early indicator of overall success.} Models that immediately recognize available tools (within 1 iteration) consistently outperform those requiring multiple iterations. Delayed recognition causes cascading failures throughout the execution pipeline. A notable planning-execution gap exists across nearly all models. \textit{Planning accuracy consistently exceeds execution accuracy}, except for \texttt{llama-405b}. This demonstrates that models can understand what needs to be done, but translating plans into correct actions remains challenging. \textit{Data integration proves particularly difficult for open-source models.} They achieve only 60-80\% accuracy in consolidation tasks, while the GPT-4 family achieves perfect performance. This underscores the complexity of synthesizing information across multiple FinOps data sources.
Perhaps most surprisingly, model size does not predict performance. \texttt{llama-405b}, with its 405B parameters, significantly underperforms the \texttt{Granite-8b} model. This suggests that domain-specific training and architectural choices matter far more than raw parameter count. These factors appear critical for specialized FinOps applications.


\section{Conclusion and Future Works}
In this study, we built a autonomous FinOps agent which attempts to simulate a real life use-case of FinOps practitioners. The agent was designed in a full-stack fashion from understanding the use-case, laying out a detailed plan consisting of sequence of steps from retrieving required data from various sources, to consolidating the data, then analyzing it to create potential recommendation for optimization. We also defined a set of metrics to evaluate the agent in different aspects like planning capability, plan execution accuracy, tool recognition, tool call accuracy, overall plan execution time. The results show that for some good LLMs, the agent achieved comparable performance as same as a real FinOps practitioner. For future work, we will expand the scope of our agent (e.g. more domain-specific knowledge, more tools) to cover more FinOps tasks.

\section{Acknowledgments}
We are thankful to Rohan Arora and Vadim Sheinin from IBM Research (US) for their valuable contributions to the brainstorming and implementation of the FinOps agent prototype.



\bibliography{aaai2026}


\clearpage
\newpage
\appendix
\section{Appendices}

\subsection{1. Examples of Execution Plan generated by FinOps Agent}
\label{appendix:plan}
\begin{figure}[h] 
    \centering 
    \includegraphics[width=1\linewidth]{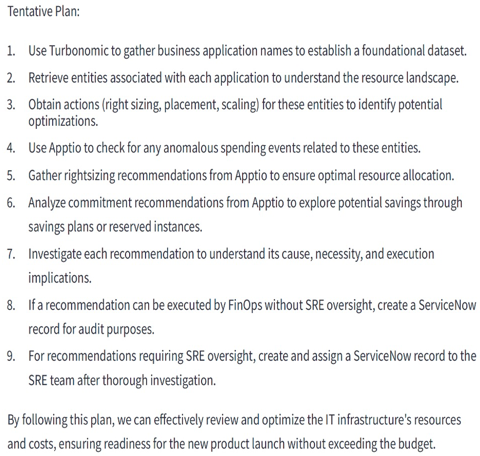}
    \caption{An Execution Plan generated by GPT-4o.}
    \label{fig:gpt-4o_plan} 
\end{figure}

Figure \ref{fig:gpt-4o_plan} shows an example of an execution plan for the use-case generated by our FinOps agent using gpt-4o model.

\begin{figure*}[h] 
    \centering 
    \includegraphics[width=1\linewidth]{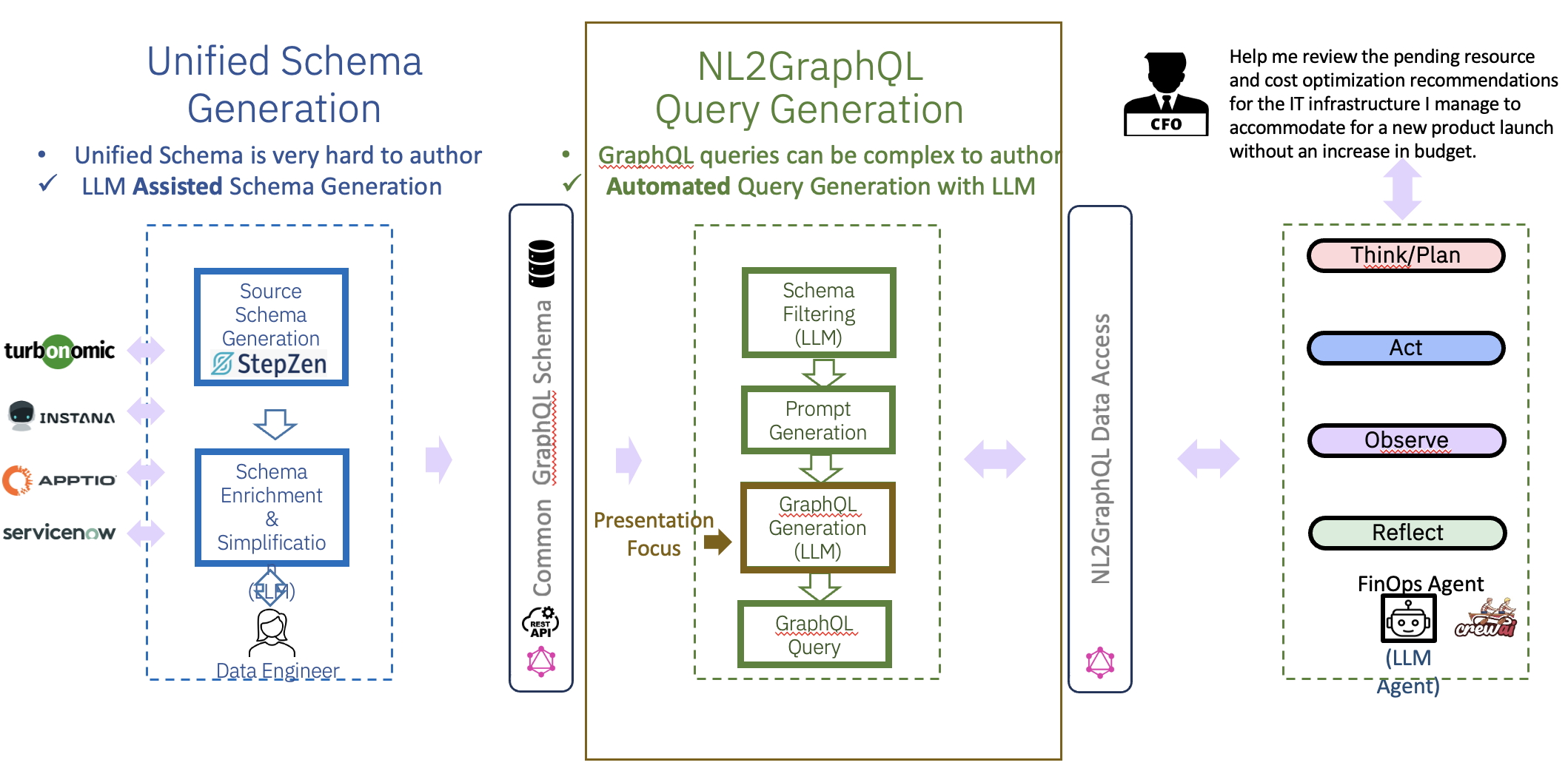}
    \caption{NL2GraphQL Architecture for FinOps Agent.}
    \label{fig:nl2graphql} 
\end{figure*}

\subsection{2. NL2GraphQL Architecture}
\label{appendix:nl2graphql}

Figure~\ref{fig:nl2graphql} illustrates the end-to-end architecture connecting the heterogeneous FinOps data sources, the unified GraphQL schema, and the NL2GraphQL query generation components with the FinOps agent’s reasoning loop. The architecture consists of three primary stages: \emph{Unified Schema Generation}, \emph{NL2GraphQL Query Generation}, and the \emph{Reasoning and Action Cycle} of the FinOps agent.

\subsubsection{Unified Schema Generation}

Constructing a unified schema across heterogeneous FinOps platforms such as IBM Turbonomic~\cite{IBMTurbonomic}, IBM Apptio~\cite{IBMApptio}, Instana, and ServiceNow is a non-trivial task. The process begins with source schema generation using StepZen \cite{stepzen}, which automatically composes schema fragments from REST endpoints, generating type definitions and query interfaces for each connected system. Next, an LLM-based schema enrichment and simplification phase harmonizes field names, merges equivalent entities (e.g., \texttt{application} in Turbonomic and \texttt{service} in Apptio), and ensures consistency in relationships and type hierarchies. This phase is executed under the supervision of a data engineer who verifies structural alignment and semantic correctness.

The result is a federated but \emph{common GraphQL schema} that encapsulates all FinOps-relevant data entities—applications, resources, costs, anomalies, and optimization recommendations—across systems. This schema serves as the unified data abstraction layer that enables cross-platform queries and forms the foundation for NL2GraphQL generation.

\subsubsection{NL2GraphQL Query Generation}

GraphQL’s expressiveness allows federated access, but authoring correct multi-source queries manually is challenging. The NL2GraphQL Query Generation component automates this process through LLM-based reasoning. It consists of three stages: \emph{Schema Filtering}, \emph{Prompt Generation}, and \emph{GraphQL Query Synthesis}.

\paragraph{Schema Filtering.} Given a natural-language request, the LLM first identifies the relevant subset of schema fields and endpoints using semantic filtering. For example, for the user query, “\emph{Review pending optimization recommendations for cost anomalies in Application~X},” the model filters schema elements related to \texttt{get\_spending\_anomaly\_events} and \texttt{get\_actions}.

\paragraph{Prompt Generation.} The system then constructs a prompt embedding both the filtered schema and the task context (e.g., user intent, entity scope, and cost constraints). This adaptive prompt ensures that the LLM’s generation is grounded in schema reality and avoids hallucinated field references.

\paragraph{GraphQL Generation.} Finally, the model generates a complete GraphQL query conforming to the schema’s structure. For example: \\

\begin{lstlisting}[language=GraphQL, caption={Example of Generated Federated GraphQL Query for FinOps Analysis}, label={lst:finops-federated}]
query ReviewOptimization {
  apptioGetSpendingAnomalyEvents(appName: "Application_X") {
    id
    anomalyType
    anomalyValue
    severity
  }
  turbonomicGetActions(appName: "Application_X") {
    id
    actionType
    risk
    recommendation
    costImpact
  }
}
\end{lstlisting}

This query federates cost anomalies from Apptio with optimization recommendations from Turbonomic, abstracting all API-level integration complexity. The generated query is validated through schema introspection before execution. This design is consistent with recent NL2GraphQL advances~\cite{kesarwani2024graphql, gupta2025schema}, where dynamic in-context learning improves query accuracy under schema constraints.

\subsubsection{FinOps Agent Reasoning Loop}

When a financial decision-maker (e.g., a CFO) issues a request such as:
\begin{quote}
\emph{“Help me review pending resource and cost optimization recommendations for our IT infrastructure to accommodate a new product launch without increasing the budget.”}
\end{quote}
the agent interprets this as a composite task involving cost anomaly detection, commitment analysis, and resource optimization. It invokes the NL2GraphQL generator to retrieve the necessary data via the unified schema, observes the retrieved outputs, and synthesizes a concise analytical summary or recommendation plan. This cycle continues until all decision requirements are satisfied.

The FinOps agent is implemented as an orchestrated LLM-based system that integrates both reasoning (natural language understanding, plan decomposition) and acting (query generation and data retrieval). The output is a structured analysis grounded in factual data retrieved through the GraphQL layer, ensuring explainability and auditability—critical in BFS and enterprise FinOps contexts.

\paragraph{Summary.}  
Together, these three components enable seamless traversal from heterogeneous financial telemetry to actionable, data-driven insights. The architecture achieves a closed loop between natural language reasoning and structured data access—bridging the gap between domain experts’ intent and multi-source infrastructure intelligence.

\subsection{3. Logic Flow of FinOps Agent for IT Infrastructure and Cost Optimization}
\label{appendix:logic}
\begin{figure*}[h] 
    \centering 
    \includegraphics[width=1.0\linewidth]{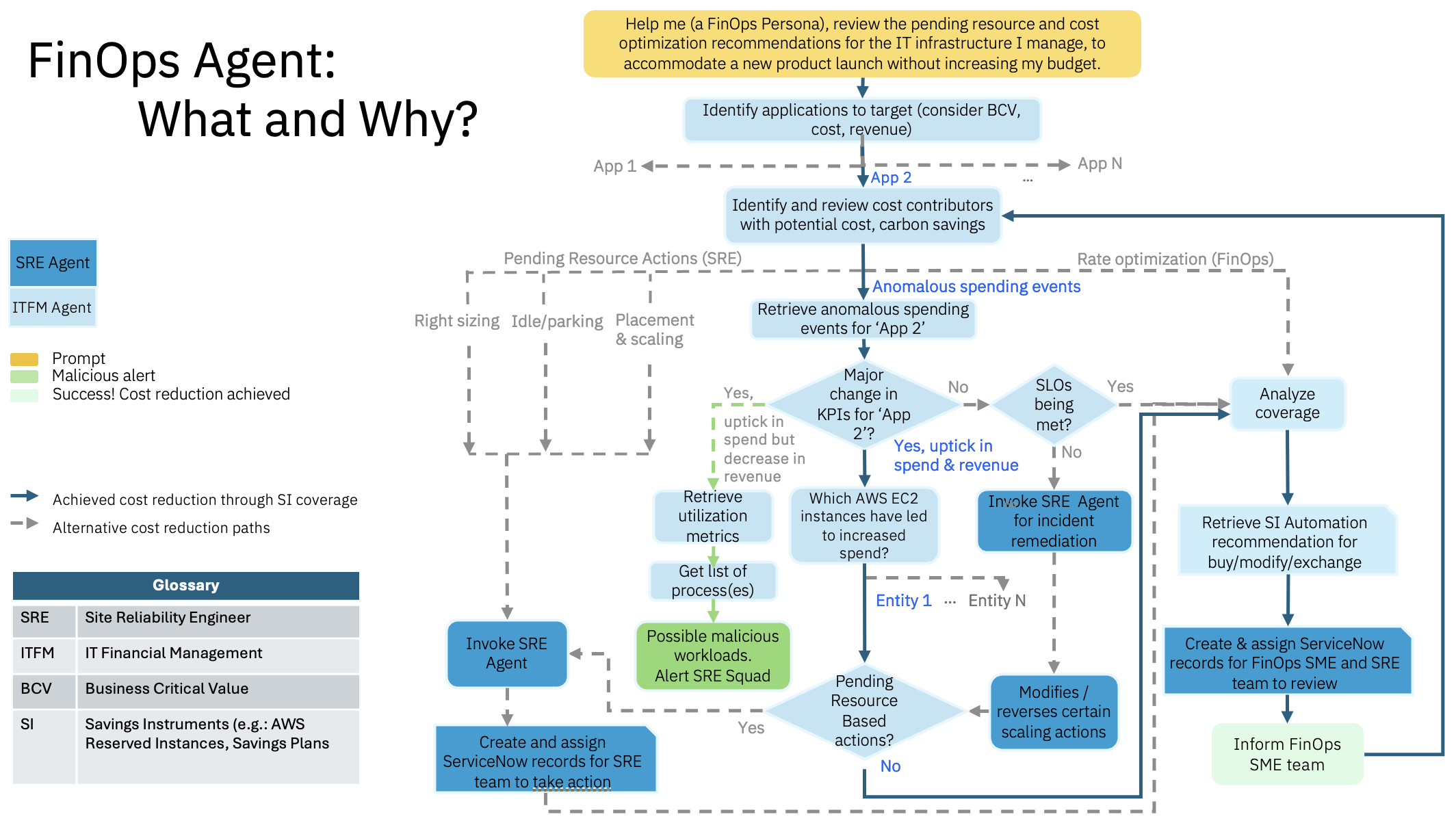}
    \caption{Logic Flow of FinOps Agent for IT Infrastructure and Cost Optimization.}
    \label{fig:finops_what-why} 
\end{figure*}

Figure \ref{fig:finops_what-why} illustrates the complex decision tree this query
generates. The agent must navigate multiple investigation
paths, with path selection contingent on data from disparate
systems.

\subsection{4. Demonstration of a Complete Run using GPT-4o}
\label{appendix:run}
We used Streamlit\footnote{https://streamlit.io/generative-ai} to build a User Interface (UI) for our FinOps agent. The following sequence of screenshots (Figure \ref{fig:run-screen1}) show a complete run of our FinOps agent (using gpt-4o) for the given use-case of IT infrastructure and cost optimization.
The run consists of the followings:
\begin{itemize}
    \item A review of given instruction and available tools
    \item A tentative plan to solve the given request
    \item Execution of sequence of steps to call data retrieval tools to obtain required data according to the tentative plan.
    \item A consolidation of all retrieved data
    \item An analysis and recommendations (as ServiceNow records) created for IT infrastructure and cost optimization.
\end{itemize}

\begin{figure*}[h] 
    \centering 
    \includegraphics[width=1.0\linewidth]{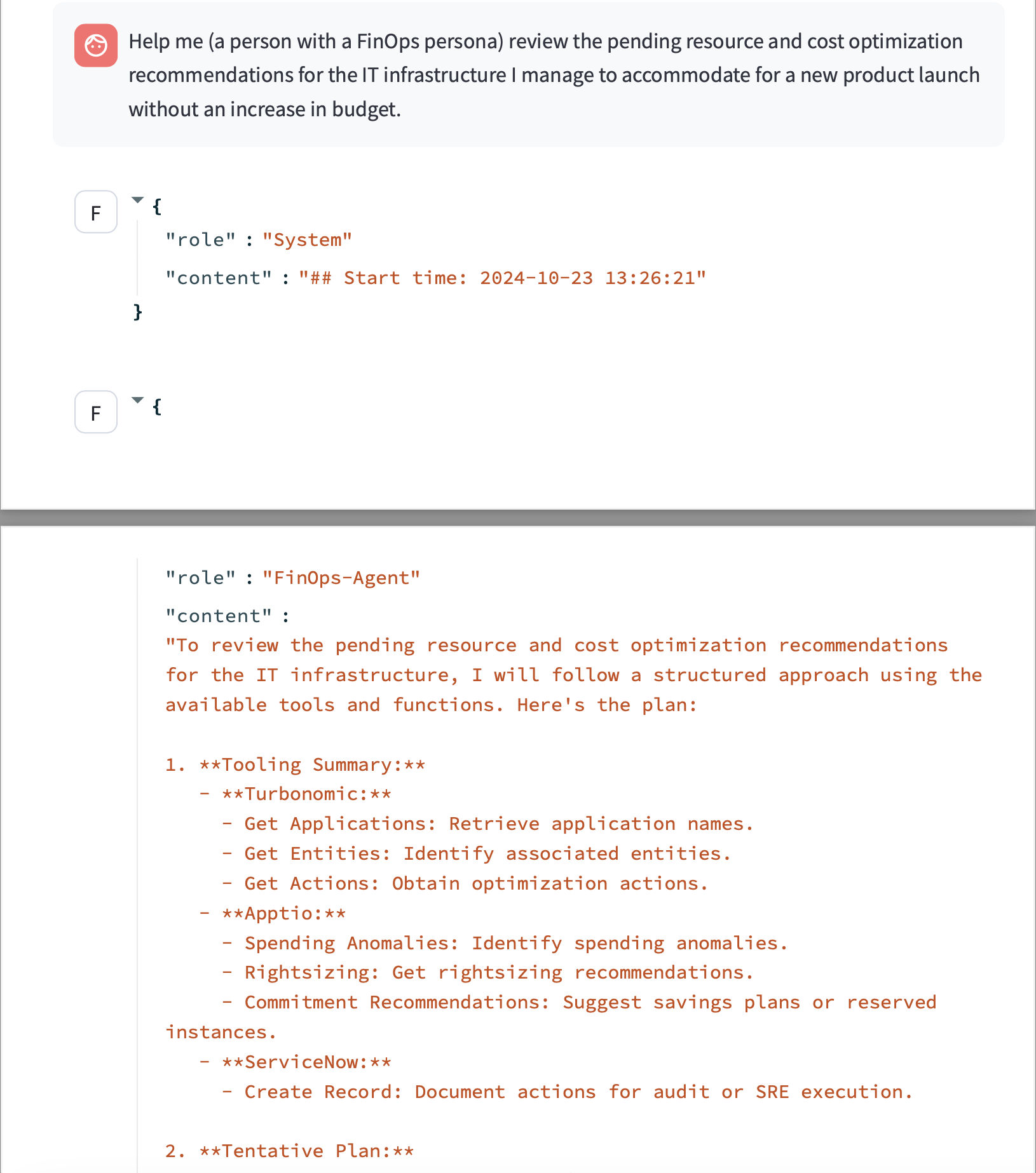}
    \label{fig:run-screen1} 
\end{figure*}

\begin{figure*}[h] 
    \centering 
    \includegraphics[width=1.0\linewidth]{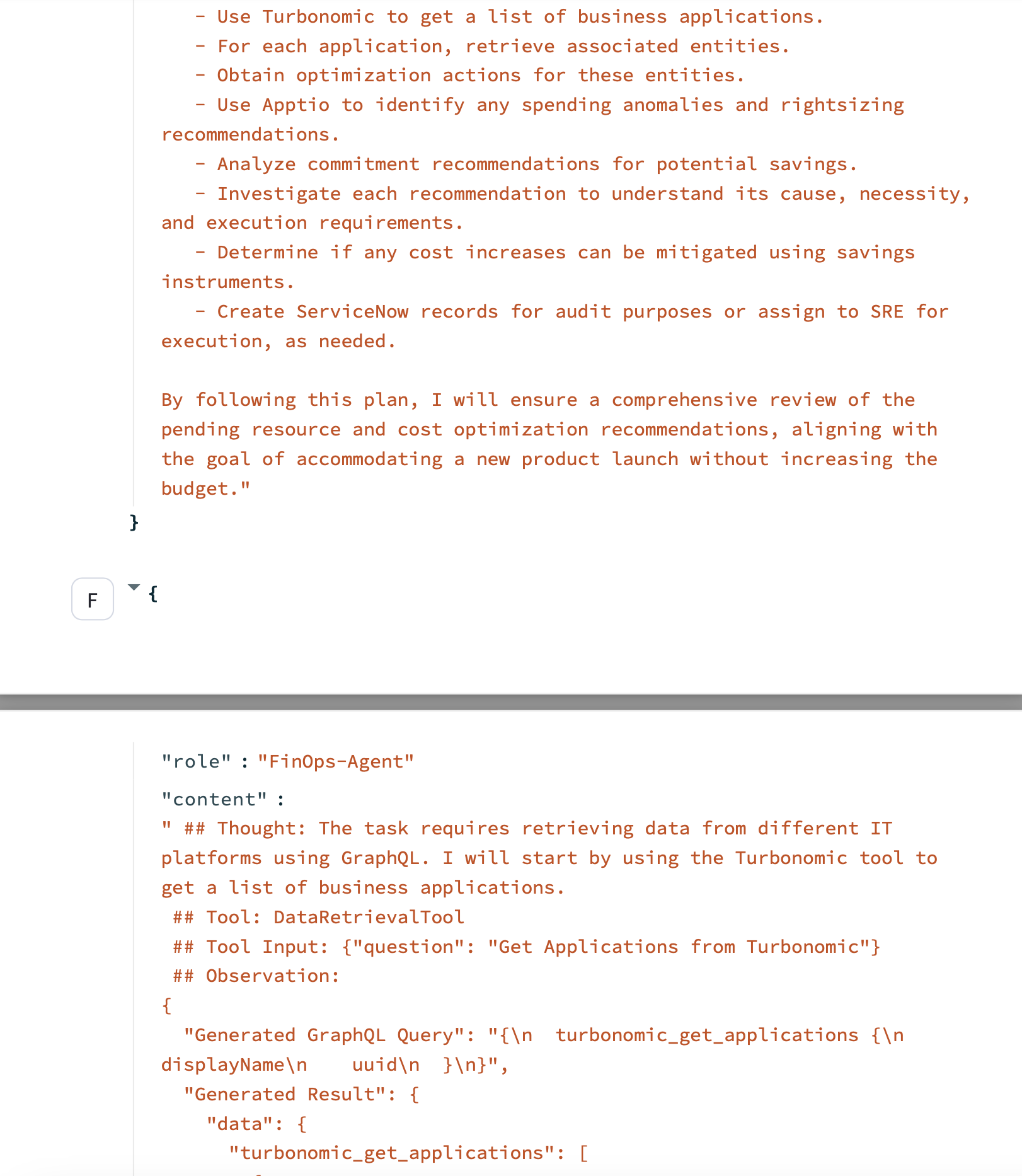}
    \label{fig:run-screen1} 
\end{figure*}

\begin{figure*}[h] 
    \centering 
    \includegraphics[width=1.0\linewidth]{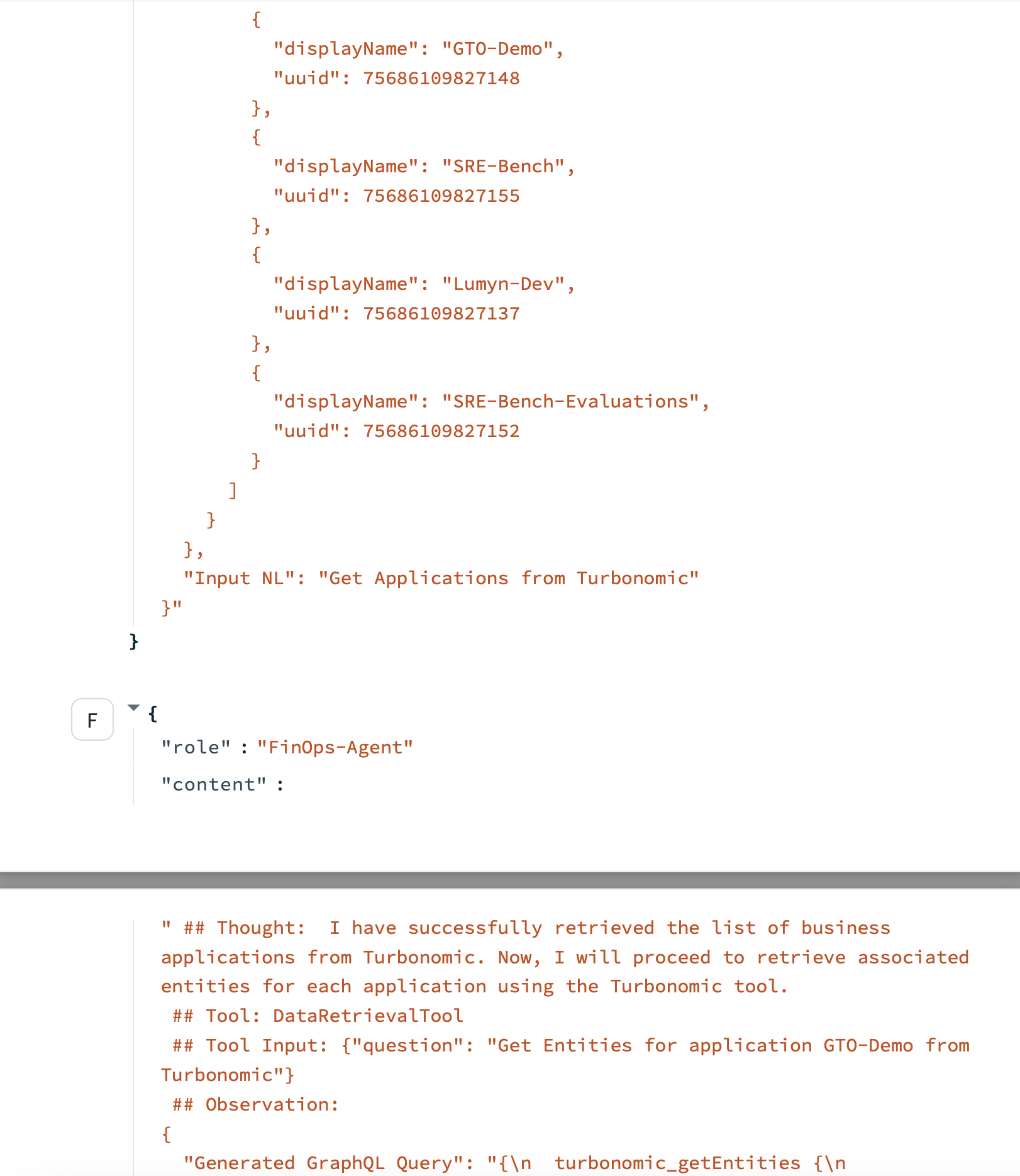}
    \label{fig:run-screen1} 
\end{figure*}

\begin{figure*}[h] 
    \centering 
    \includegraphics[width=1.0\linewidth]{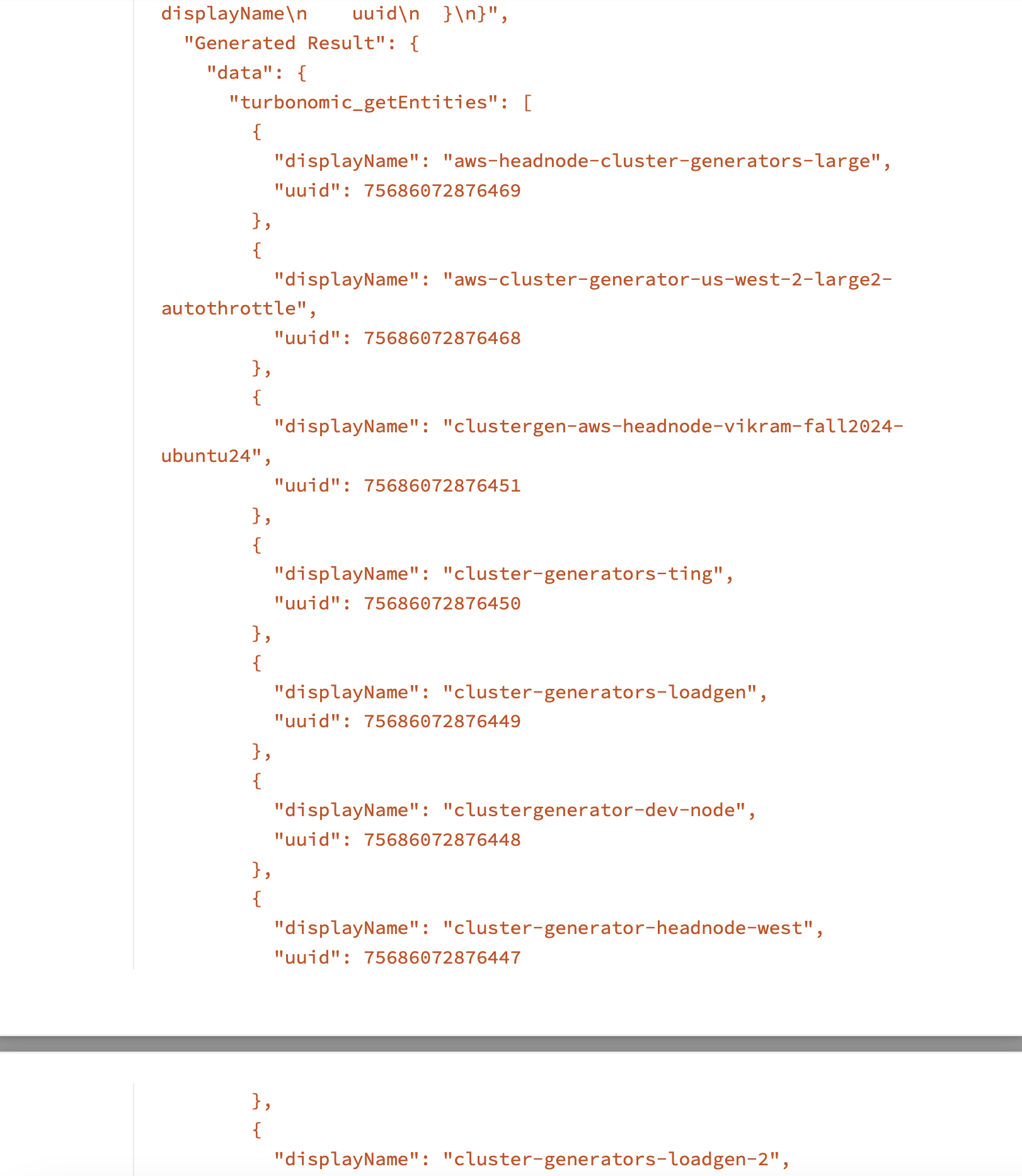}
    \label{fig:run-screen1} 
\end{figure*}

\begin{figure*}[h] 
    \centering 
    \includegraphics[width=1.0\linewidth]{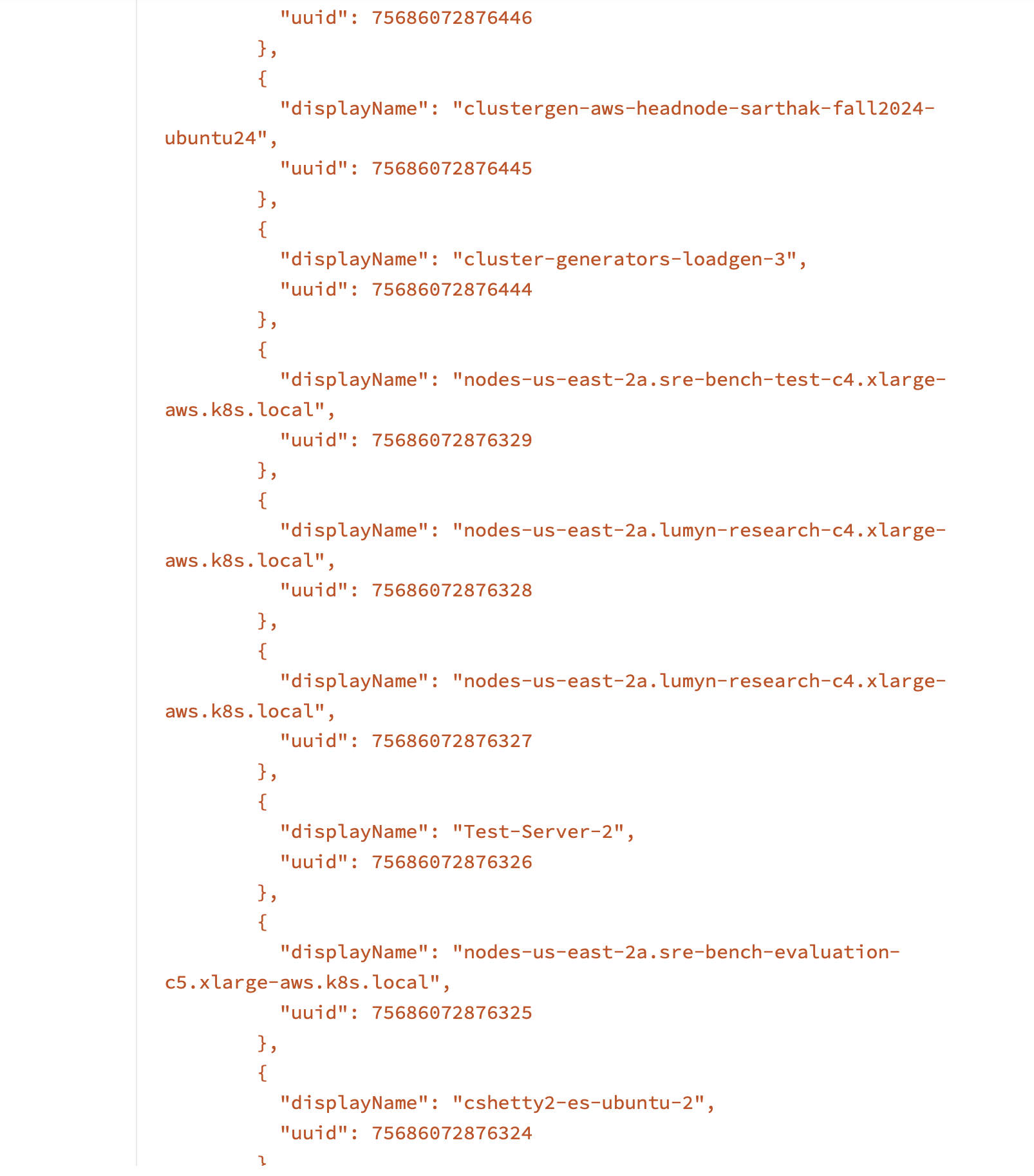}
    \label{fig:run-screen1} 
\end{figure*}

\begin{figure*}[h] 
    \centering 
    \includegraphics[width=1.0\linewidth]{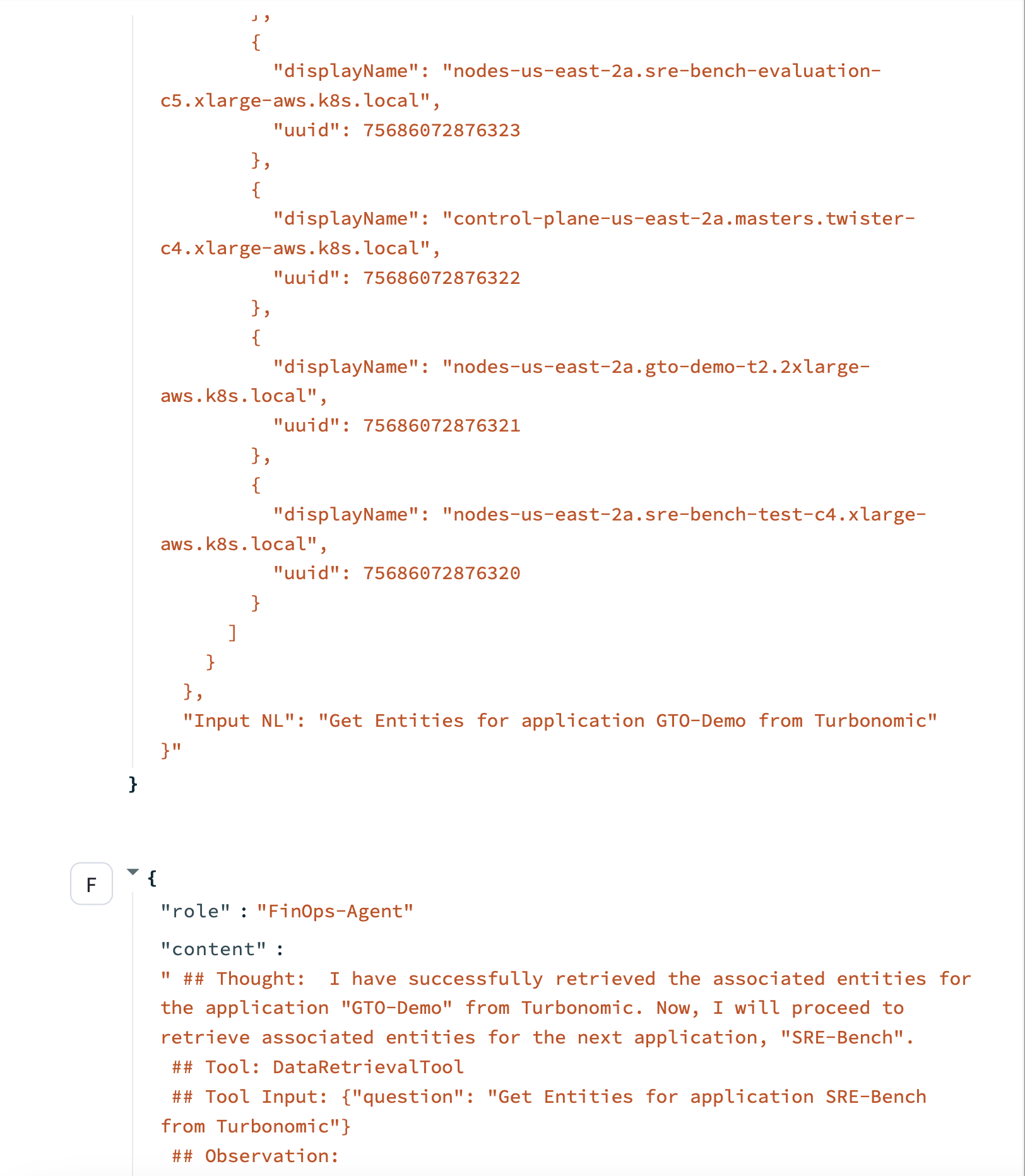}
    \label{fig:run-screen1} 
\end{figure*}

\begin{figure*}[h] 
    \centering 
    \includegraphics[width=1.0\linewidth]{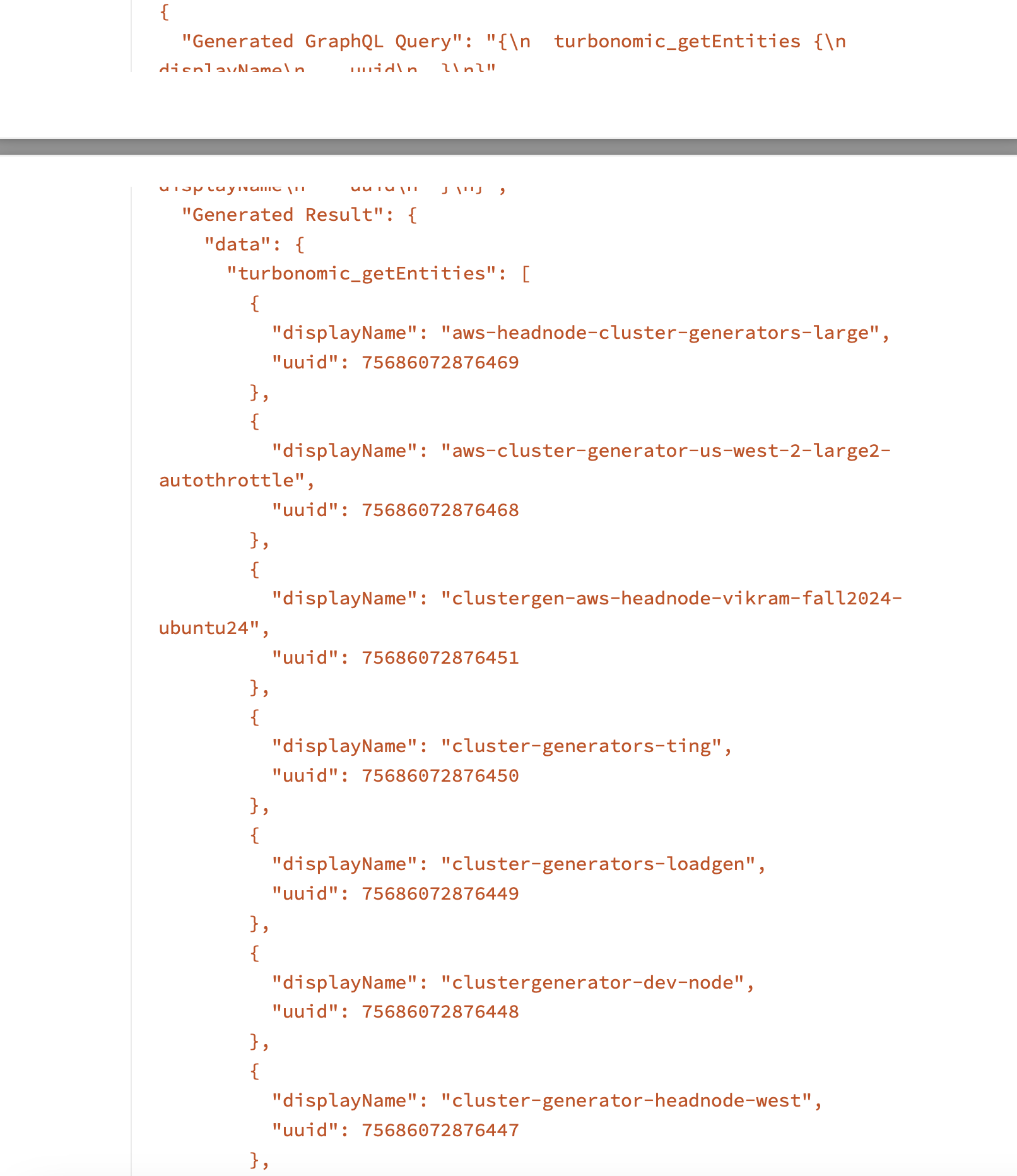}
    \label{fig:run-screen1} 
\end{figure*}

\begin{figure*}[h] 
    \centering 
    \includegraphics[width=1.0\linewidth]{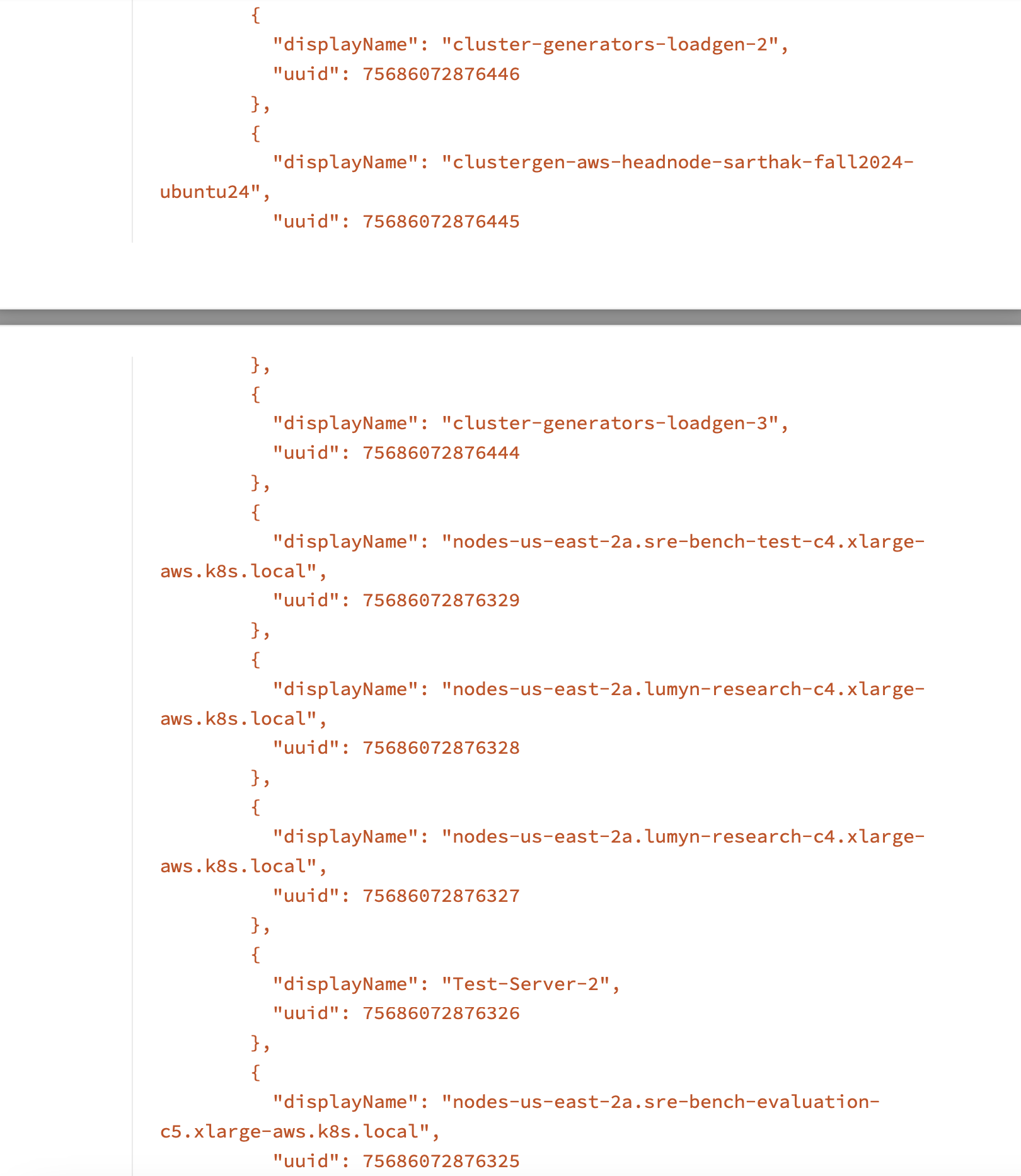}
    \label{fig:run-screen1} 
\end{figure*}

\begin{figure*}[h] 
    \centering 
    \includegraphics[width=1.0\linewidth]{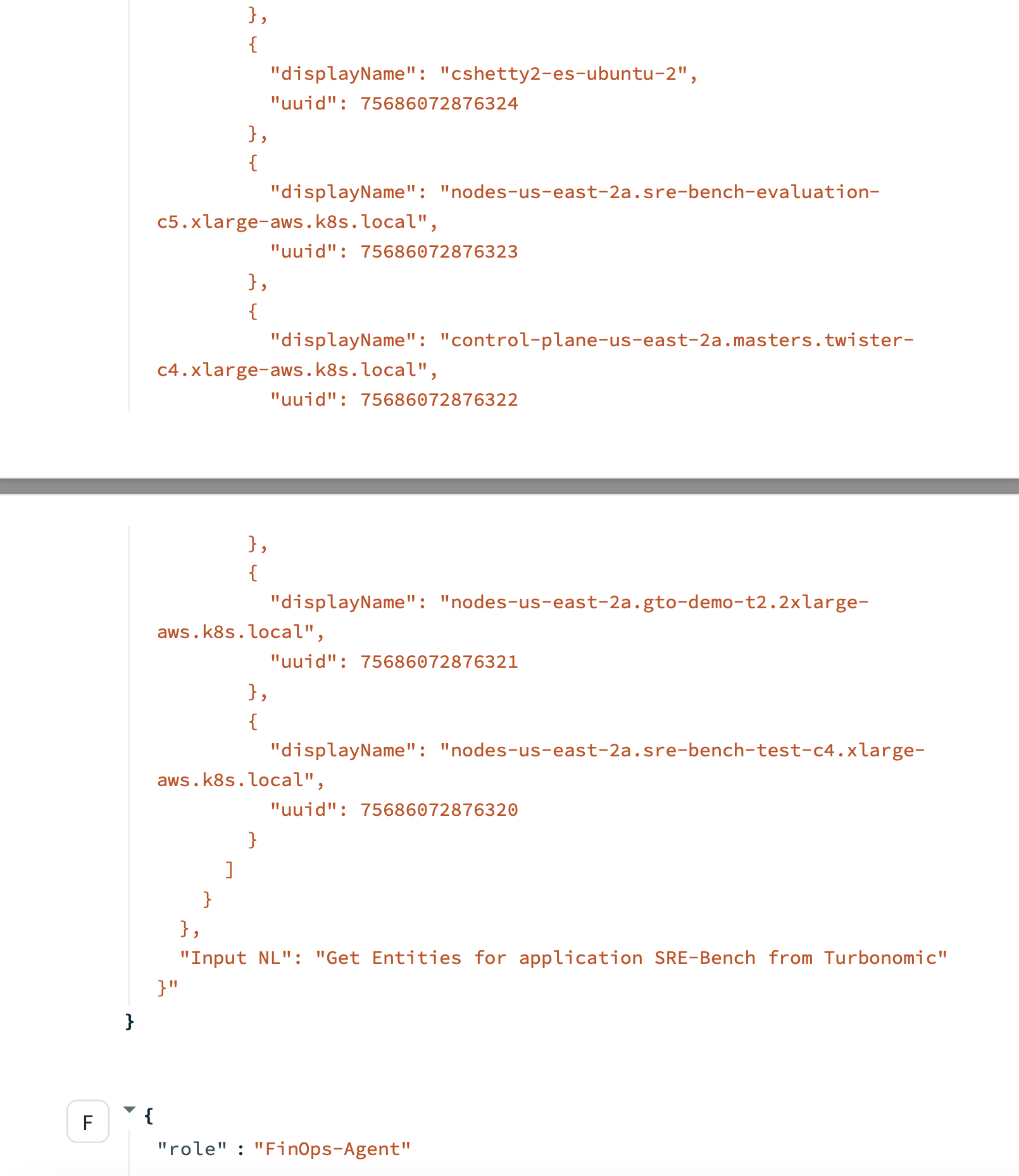}
    \label{fig:run-screen1} 
\end{figure*}

\begin{figure*}[h] 
    \centering 
    \includegraphics[width=1.0\linewidth]{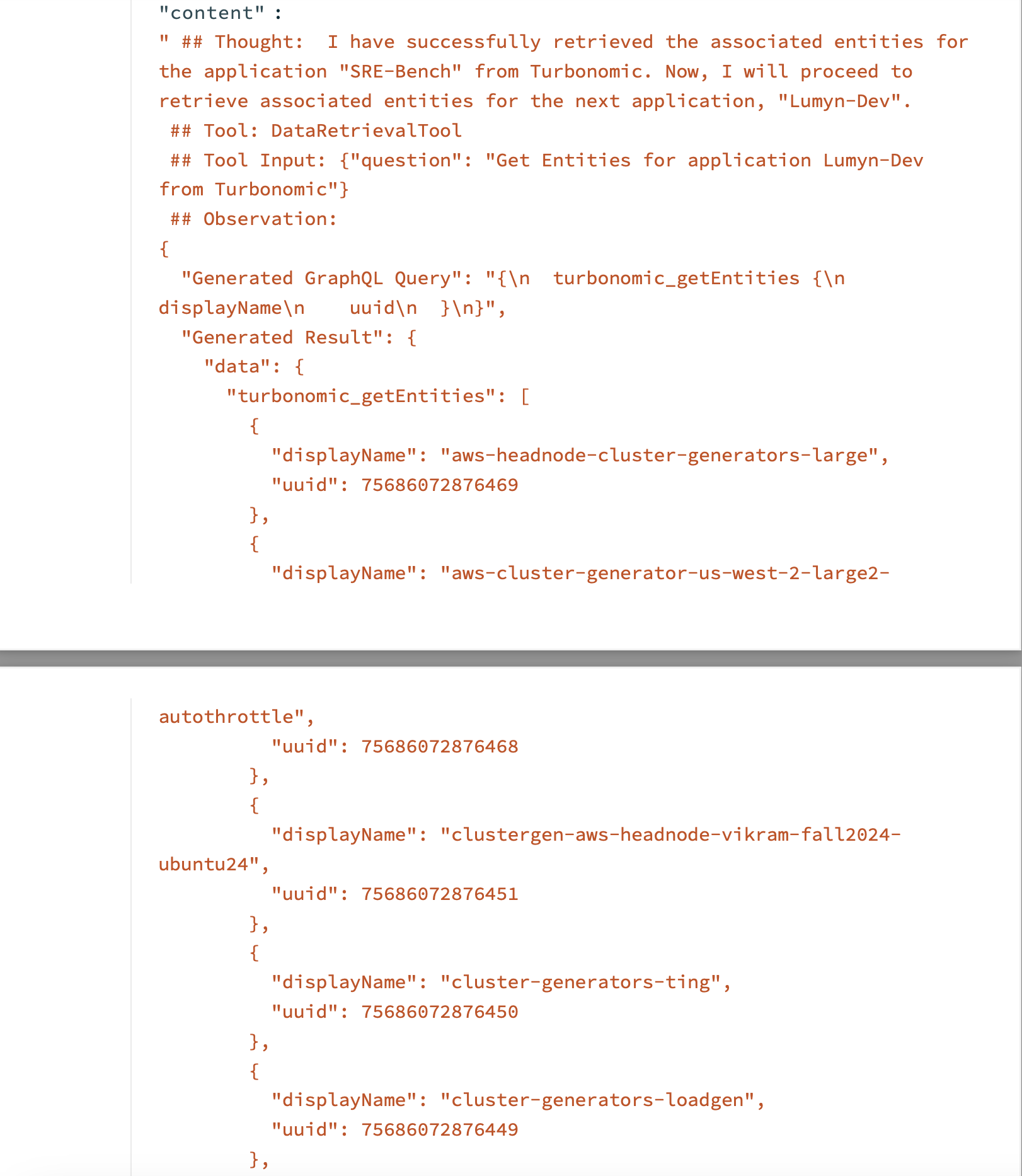}
    \label{fig:run-screen1} 
\end{figure*}

\begin{figure*}[h] 
    \centering 
    \includegraphics[width=1.0\linewidth]{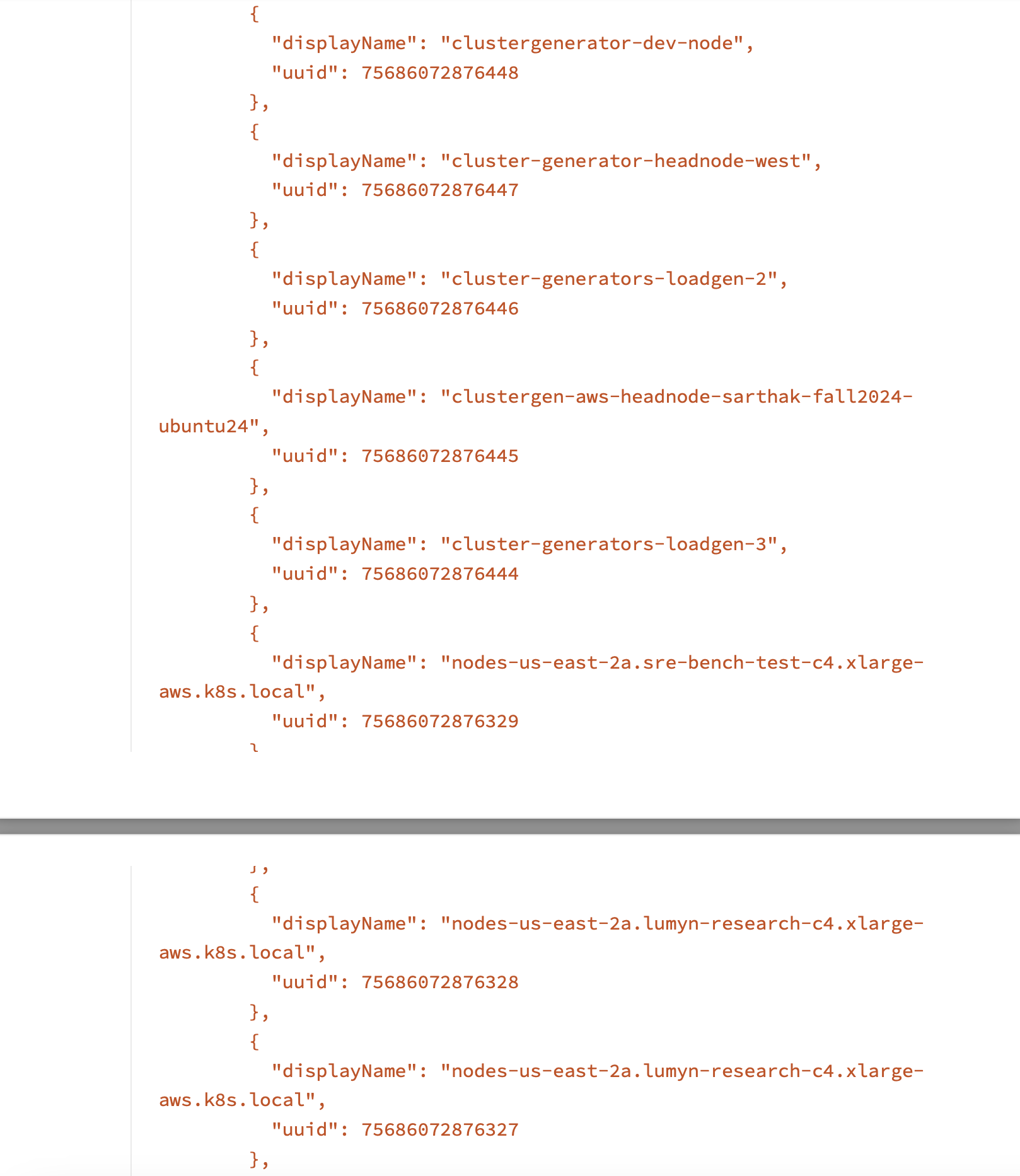}
    \label{fig:run-screen1} 
\end{figure*}

\begin{figure*}[h] 
    \centering 
    \includegraphics[width=1.0\linewidth]{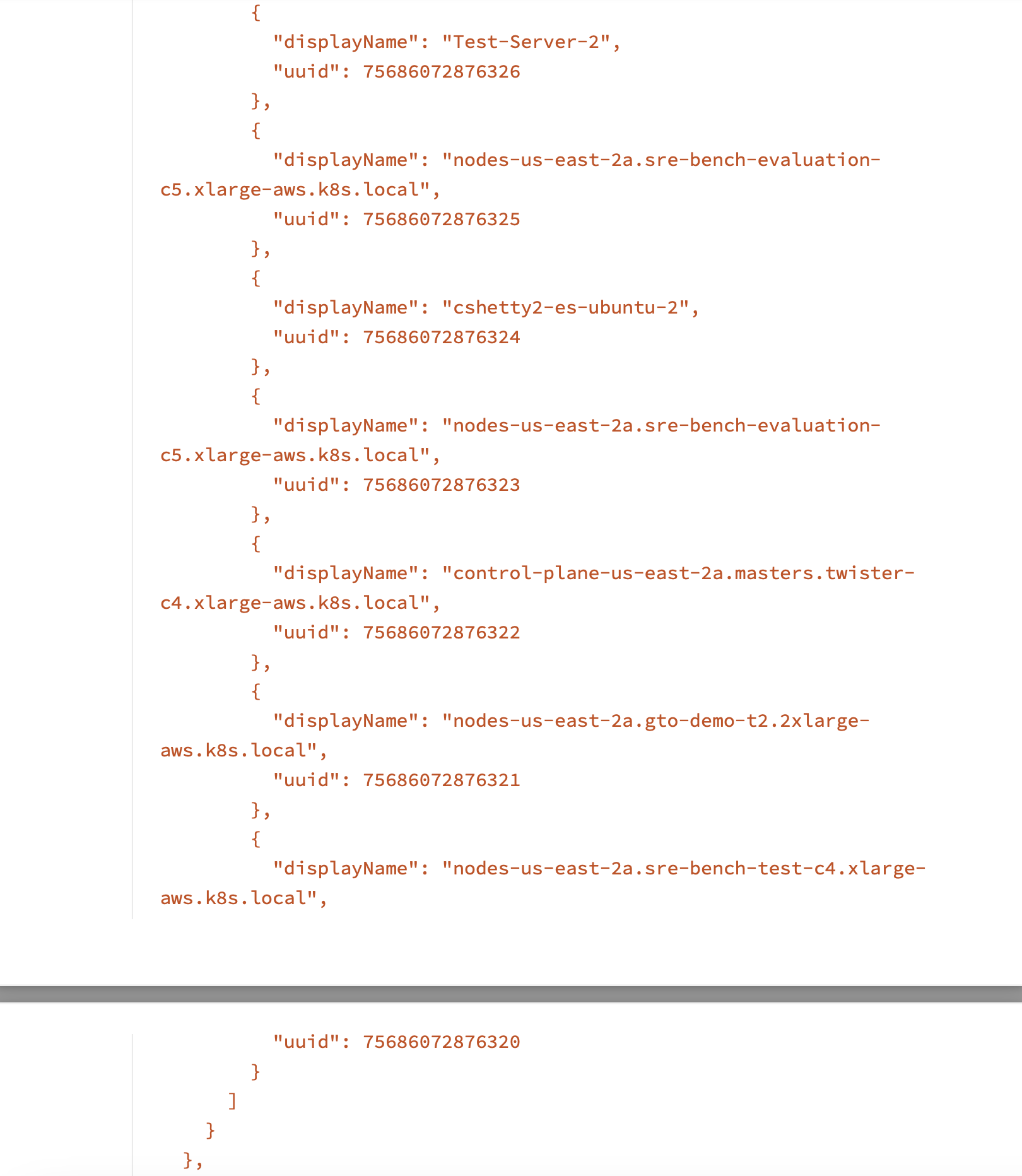}
    \label{fig:run-screen1} 
\end{figure*}

\begin{figure*}[h] 
    \centering 
    \includegraphics[width=1.0\linewidth]{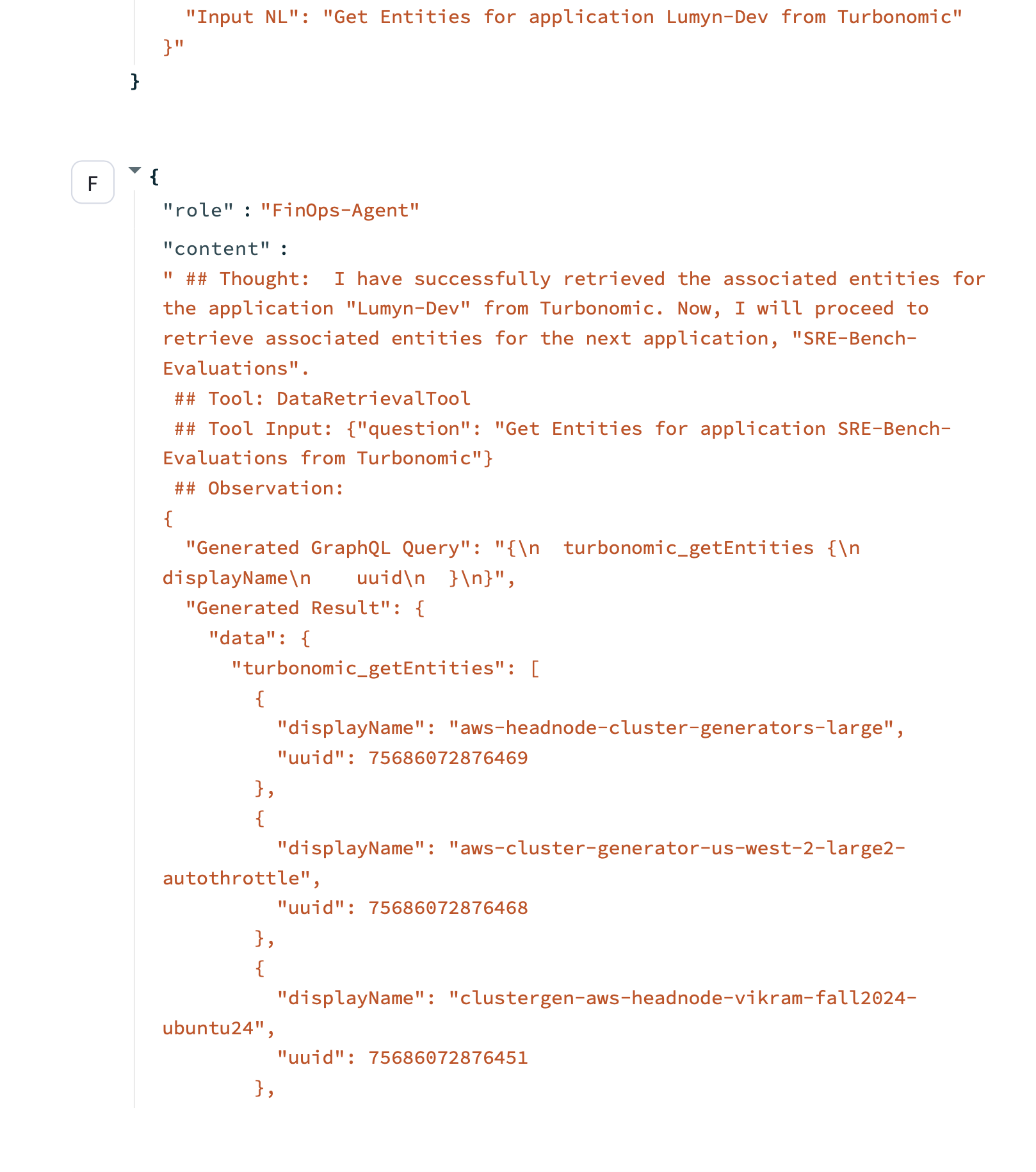}
    \label{fig:run-screen1} 
\end{figure*}

\begin{figure*}[h] 
    \centering 
    \includegraphics[width=1.0\linewidth]{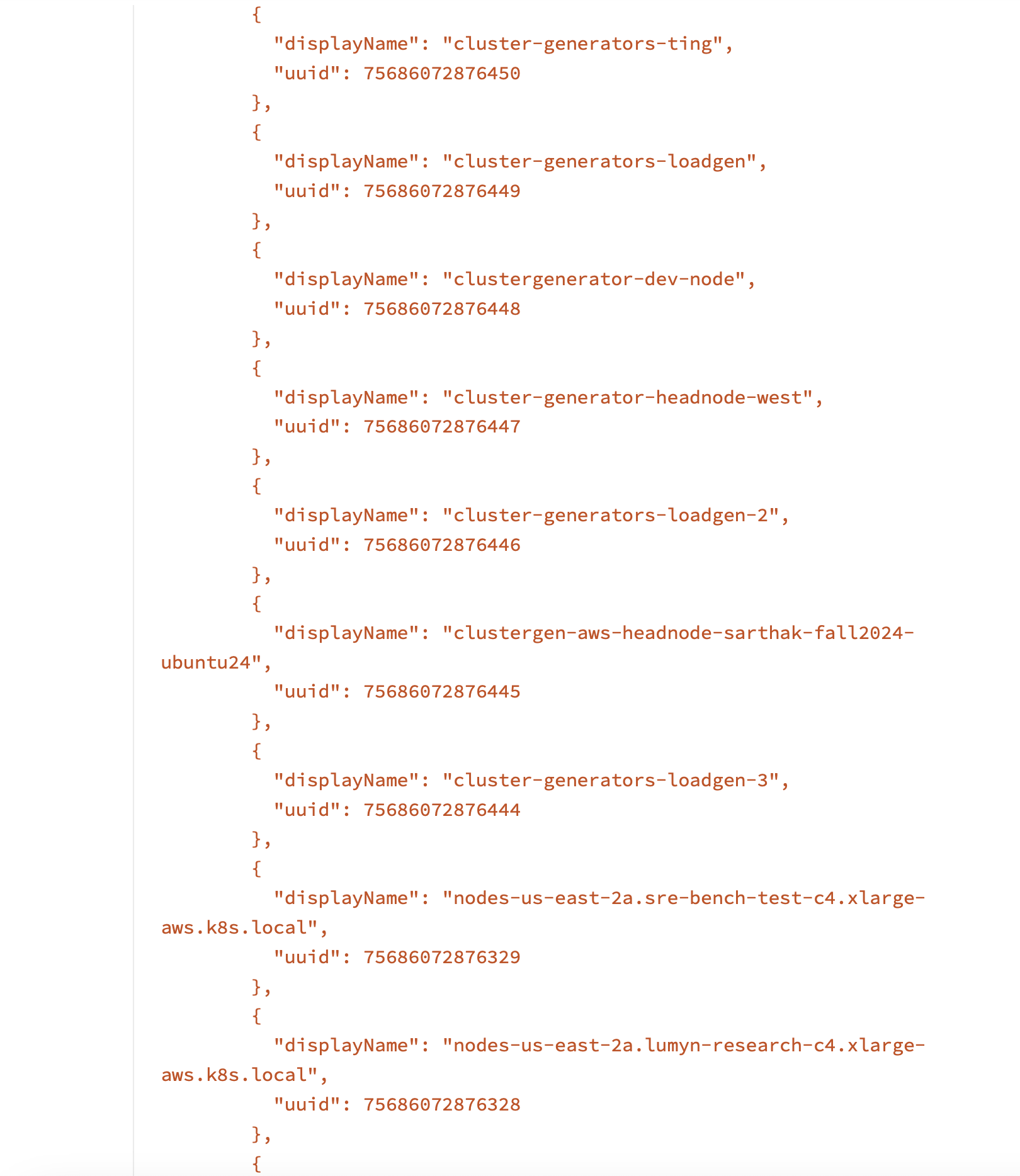}
    \label{fig:run-screen1} 
\end{figure*}

\begin{figure*}[h] 
    \centering 
    \includegraphics[width=1.0\linewidth]{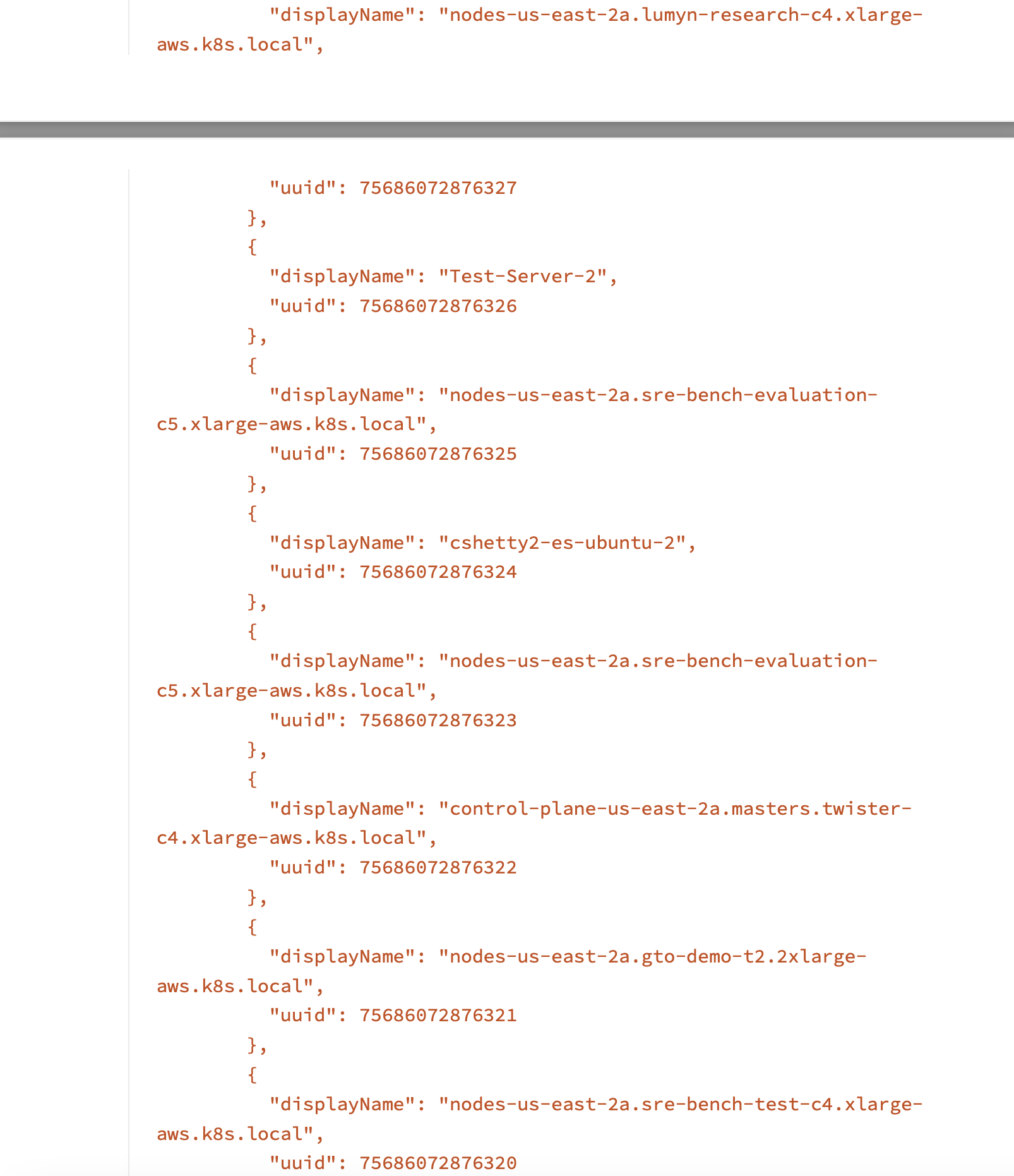}
    \label{fig:run-screen1} 
\end{figure*}

\begin{figure*}[h] 
    \centering 
    \includegraphics[width=1.0\linewidth]{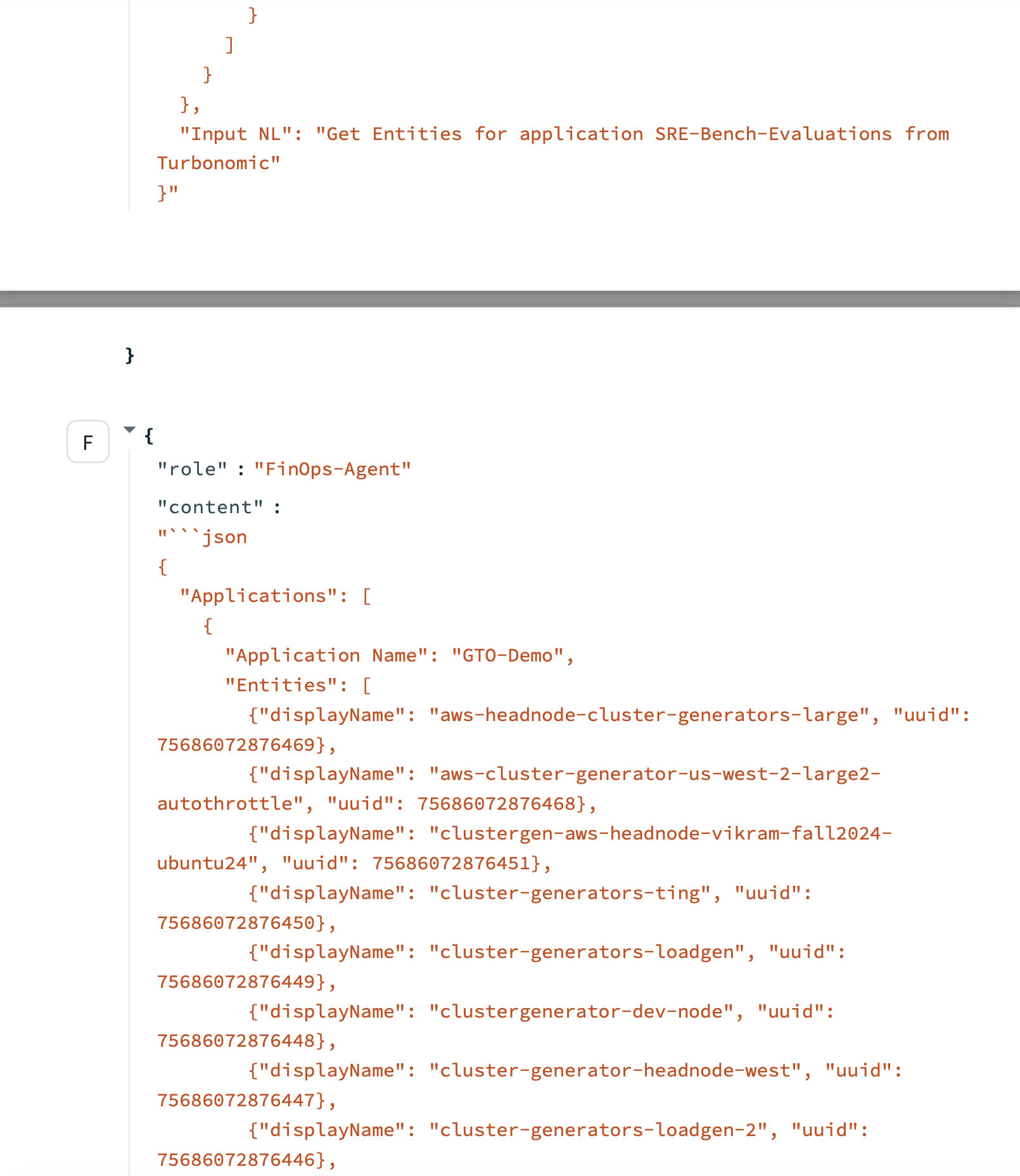}
    \label{fig:run-screen1} 
\end{figure*}

\begin{figure*}[h] 
    \centering 
    \includegraphics[width=1.0\linewidth]{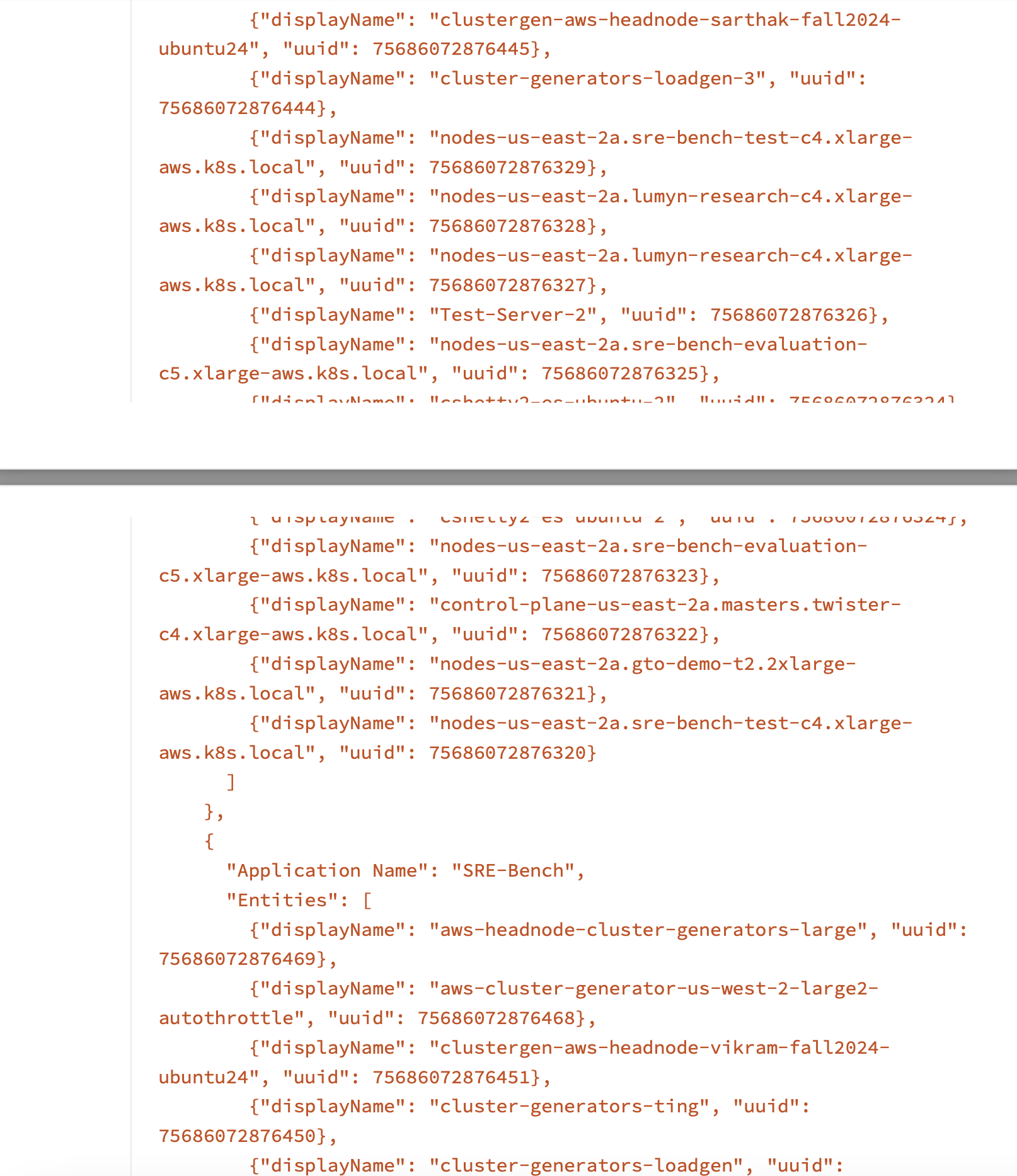}
    \label{fig:run-screen1} 
\end{figure*}

\begin{figure*}[h] 
    \centering 
    \includegraphics[width=1.0\linewidth]{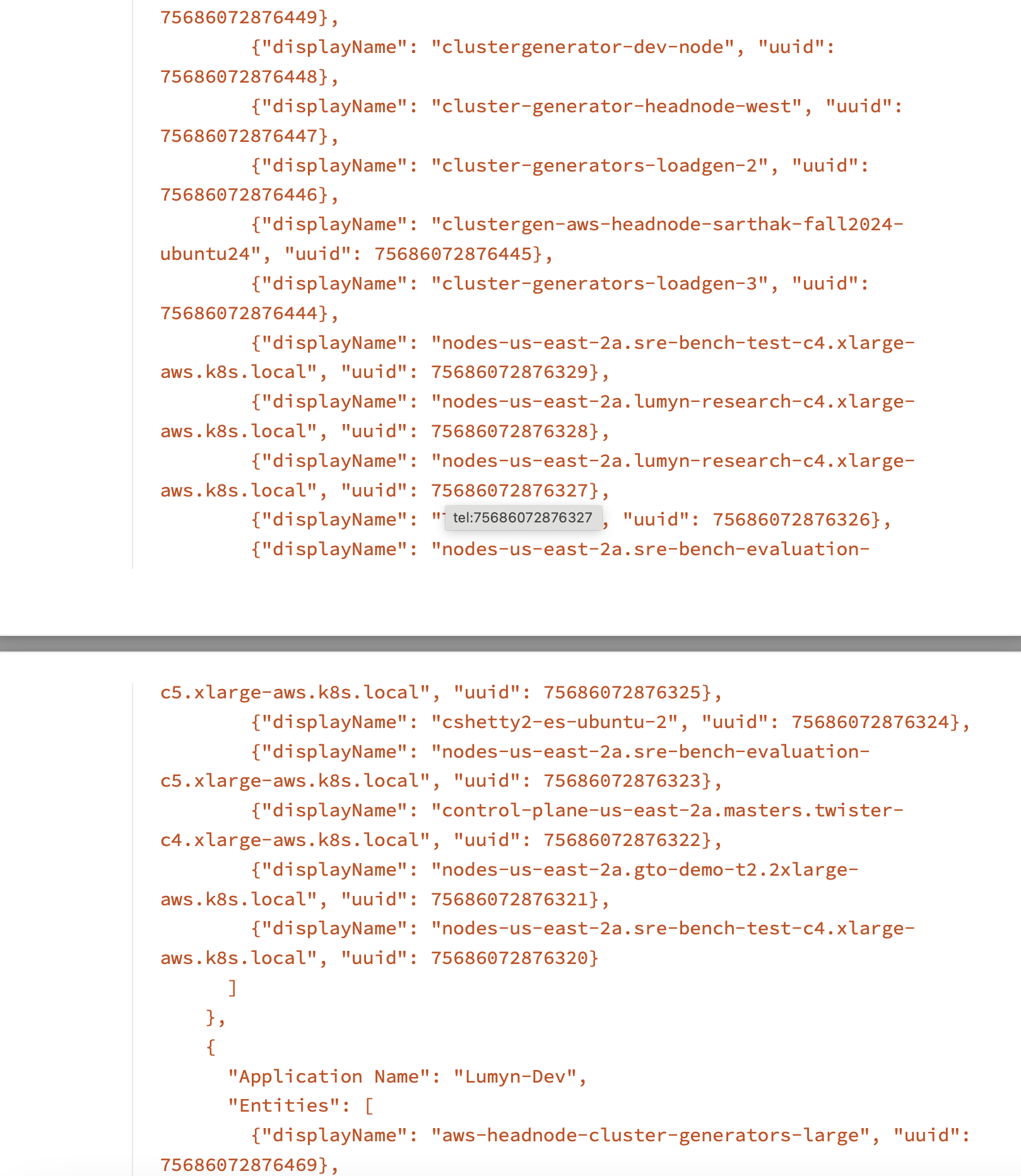}
    \label{fig:run-screen1} 
\end{figure*}

\begin{figure*}[h] 
    \centering 
    \includegraphics[width=1.0\linewidth]{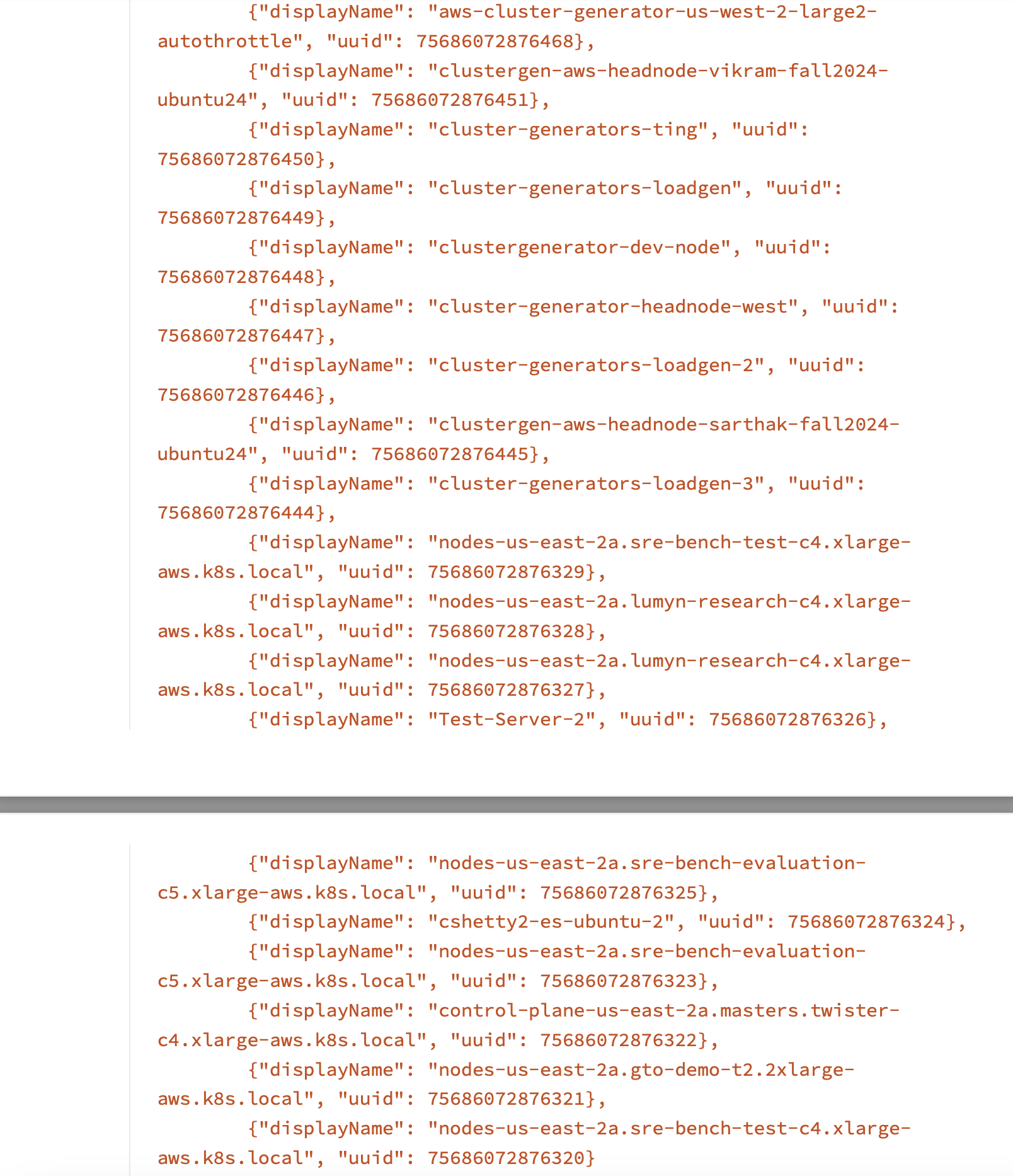}
    \label{fig:run-screen1} 
\end{figure*}

\begin{figure*}[h] 
    \centering 
    \includegraphics[width=1.0\linewidth]{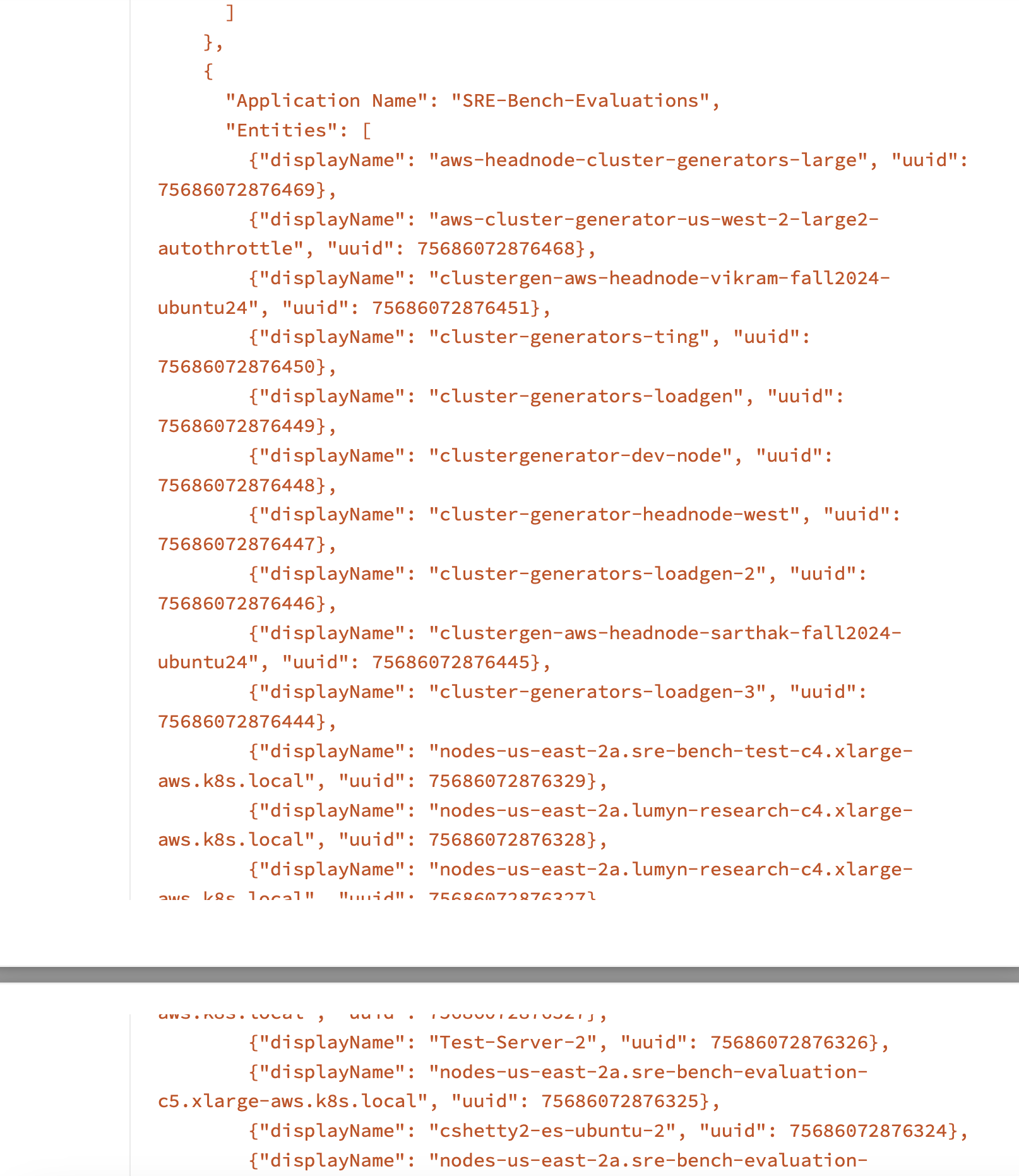}
    \label{fig:run-screen1} 
\end{figure*}

\begin{figure*}[h] 
    \centering 
    \includegraphics[width=1.0\linewidth]{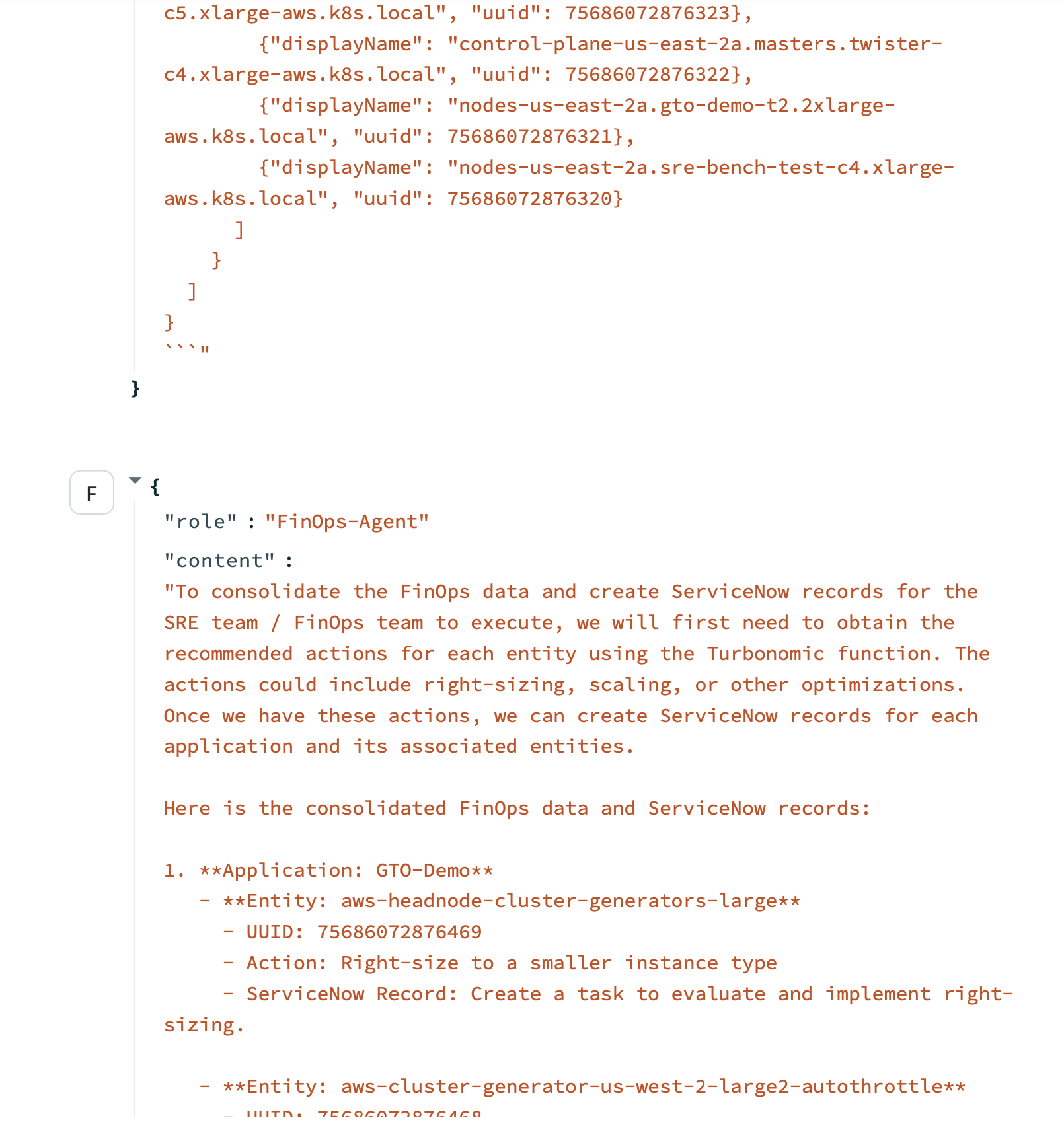}
    \label{fig:run-screen1} 
\end{figure*}

\begin{figure*}[h] 
    \centering 
    \includegraphics[width=1.0\linewidth]{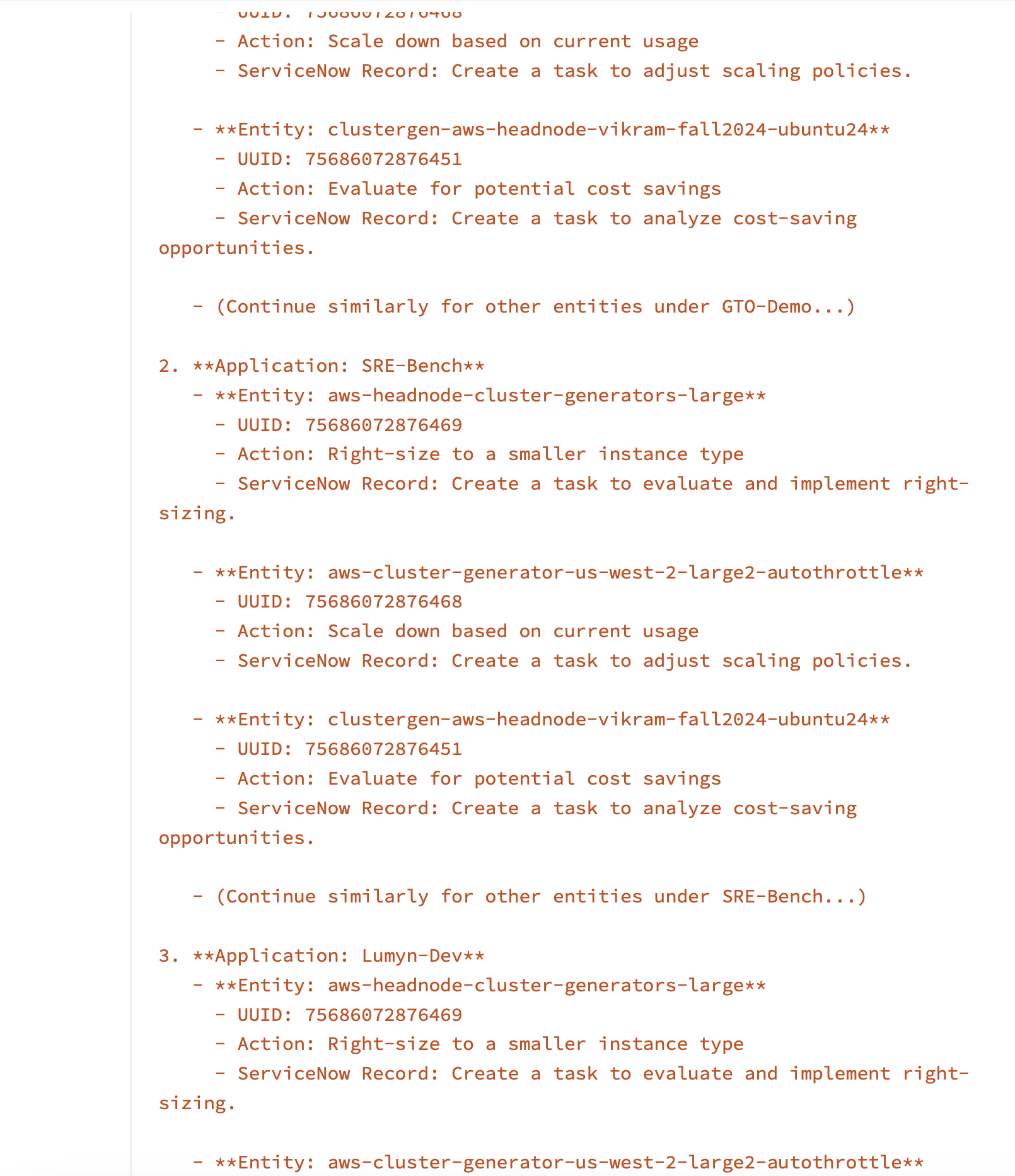}
    \label{fig:run-screen1} 
\end{figure*}

\begin{figure*}[h] 
    \centering 
    \includegraphics[width=1.0\linewidth]{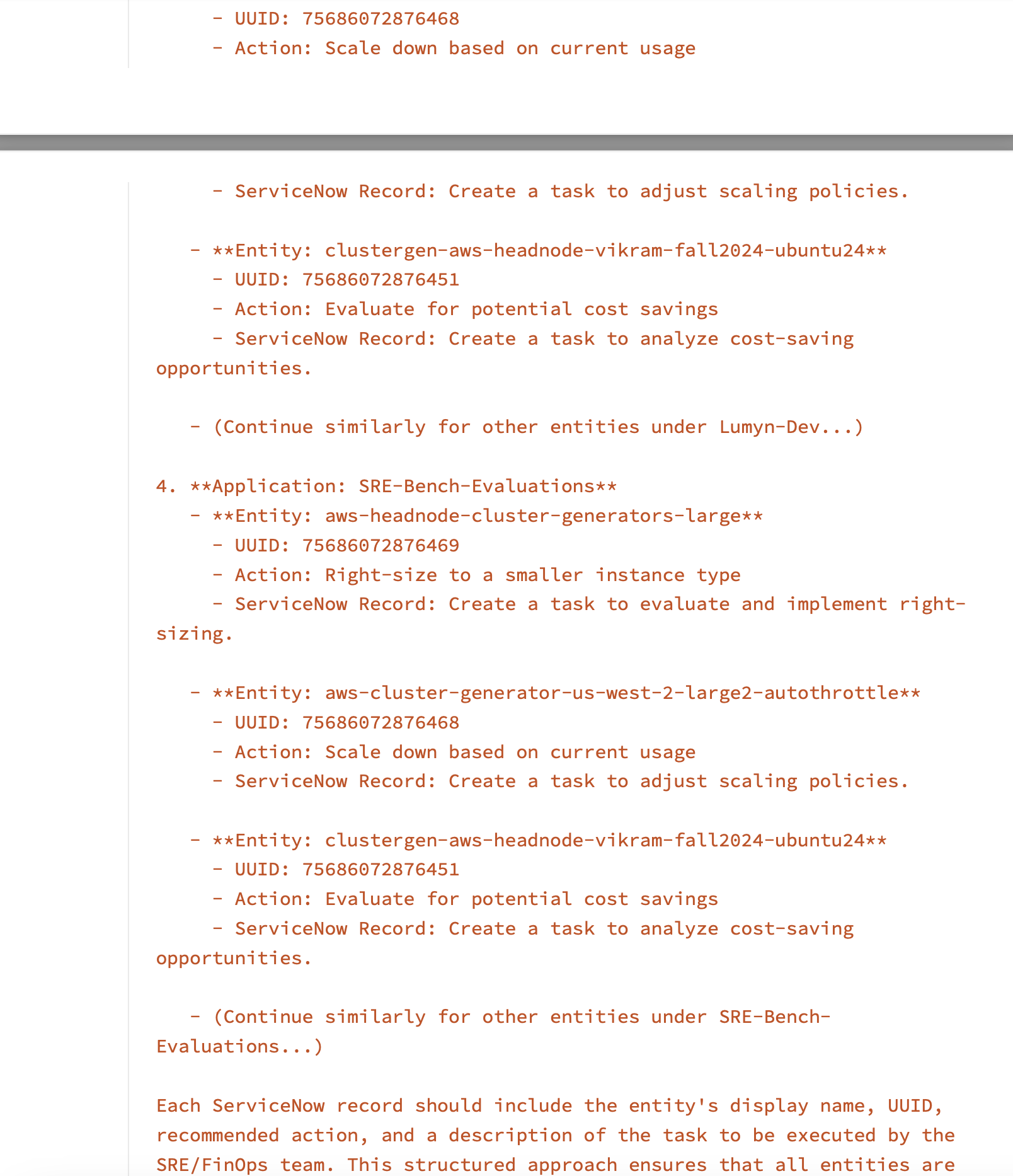}
    \label{fig:run-screen1} 
\end{figure*}

\begin{figure*}[h] 
    \centering 
    \includegraphics[width=1.0\linewidth]{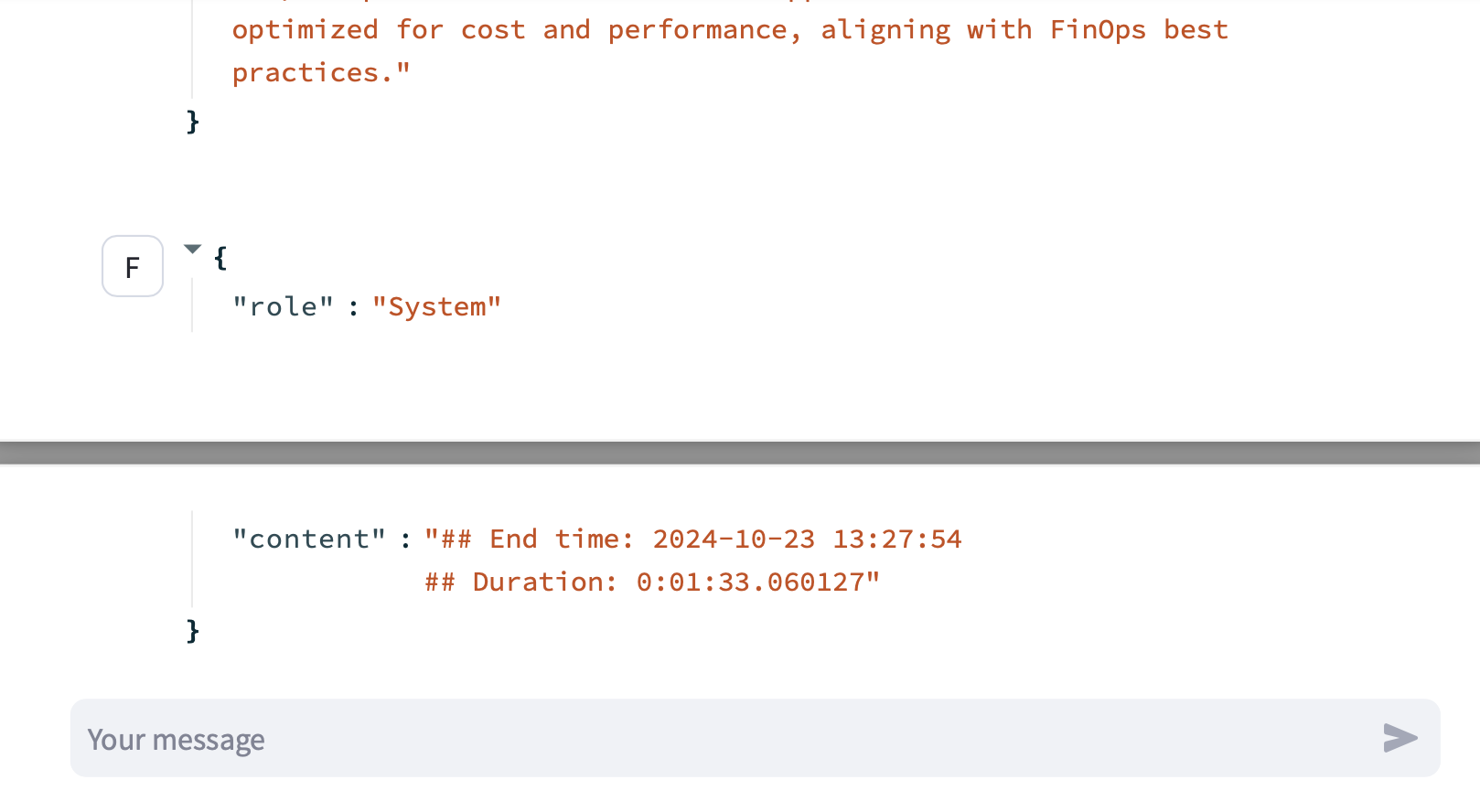}
    \caption{Demonstration of a Complete Run of FinOps Agent.}
    \label{fig:run-screen1} 
\end{figure*}

\end{document}